\documentclass[10pt,twocolumn,letterpaper]{article}

\RequirePackage[absolute]{textpos}
\DeclareRobustCommand*{\copyrightnote}{%
  \begin{textblock}{85}(17.5,259.5)
      \scriptsize{\noindent \copyright 2022 IEEE. Personal use of this material is permitted. Permission from IEEE must be obtained for all other uses, in any current or future media, including reprinting/republishing this material for advertising or promotional purposes, creating new collective works, for resale or redistribution to servers or lists, or reuse of any copyrighted
component of this work in other works.}%
  \end{textblock}%
    }

\usepackage[pagenumbers]{cvpr} 

\usepackage{graphicx}
\usepackage{amsmath}
\usepackage{amssymb}
\usepackage{booktabs}
\usepackage{makecell}
\usepackage{multirow}
\usepackage{float}
\usepackage{diagbox}
\usepackage{algorithm}
\usepackage{algpseudocode}
\usepackage[noend]{algcompatible}
\usepackage{algorithmicx}
\usepackage{xcolor}
\usepackage{adjustbox}
\usepackage{paralist}
\usepackage{enumitem}
\usepackage{mdwlist}

\usepackage{color}

\usepackage{lipsum}

\newcommand\blfootnote[1]{%
  \begingroup
  \renewcommand\thefootnote{}\footnote{#1}%
  \addtocounter{footnote}{-1}%
  \endgroup
}

 \algnewcommand\algorithmicreturn{\textbf{return}}
 \algnewcommand\RETURN{\State \algorithmicreturn}%

\newcommand{\StateRed}[1]{\algrenewcommand{\alglinenumber}[1]{\footnotesize\textcolor{red}{##1}:}\State #1}
\newcommand{\StateBlack}[1]{\algrenewcommand{\alglinenumber}[1]{\footnotesize##1:}\State #1}

\DeclareMathOperator*{\argmax}{arg\,max}

\definecolor{aquamarine}{RGB}{127, 255, 212}
\definecolor{inchworm}{rgb}{0.7, 0.93, 0.36}
\definecolor{lightfuchsiapink}{rgb}{0.98, 0.52, 0.9}
\definecolor{babyblue}{rgb}{0.54, 0.81, 0.94}
\definecolor{ccolor}{rgb}{0.74, 0.83, 0.9} 
\definecolor{scolor}{rgb}{0.53, 0.66, 0.42}

\DeclareMathOperator{\LID}{LID}

\newlength\myboxwidth

\usepackage[pagebackref,breaklinks,colorlinks]{hyperref}

\usepackage[capitalize]{cleveref}
\crefname{section}{Sec.}{Secs.}
\Crefname{section}{Section}{Sections}
\Crefname{table}{Table}{Tables}
\crefname{table}{Tab.}{Tabs.}

\def\cvprPaperID{8257} 
\def\confName{CVPR}

\begin{document}

\title{FedCorr: Multi-Stage Federated Learning for  Label Noise Correction}

\author{Jingyi Xu$^{1*}$
\qquad Zihan Chen$^{1,2}$\thanks{Equal contributions. $^\dag$ Corresponding author.}
\qquad Tony Q.S. Quek$^{1}$
\qquad Kai Fong Ernest Chong$^{1\dag}$
\\
$^1$Singapore University of Technology and Design
\quad
$^2$National University of Singapore\\
{\tt\small \{jinyi\_xu,zihan\_chen\}@mymail.sutd.edu.sg}
\quad
{\tt\small \{tonyquek,ernest\_chong\}@sutd.edu.sg}
}
\maketitle
\copyrightnote

\begin{abstract}
   Federated learning (FL) is a privacy-preserving distributed learning paradigm that enables clients to jointly train a global model. In real-world FL implementations, client data could have label noise, and different clients could have vastly different label noise levels. Although there exist methods in centralized learning  for tackling label noise, such methods do not perform well on heterogeneous label noise in FL settings, due to the typically smaller sizes of client datasets and data privacy requirements in FL. In this paper, we propose \texttt{FedCorr}, a general multi-stage framework to tackle heterogeneous label noise in FL, without making any assumptions on the noise models of local clients, while still maintaining client data privacy. In particular, (1) \texttt{FedCorr} dynamically identifies noisy clients by exploiting the dimensionalities of the model prediction subspaces  independently measured on all clients, and then identifies incorrect labels on noisy clients based on per-sample losses. To deal with data heterogeneity and to increase  training stability, we propose an adaptive local proximal regularization term that is based on estimated local noise levels. (2) We further finetune the global model on identified clean clients and correct the noisy labels for the remaining noisy clients after finetuning. (3) Finally, we apply the usual training on all clients to make full use of all local data. Experiments conducted on CIFAR-10/100 with federated synthetic label noise, and on a real-world noisy dataset, Clothing1M, demonstrate that \texttt{FedCorr} is robust to label noise and substantially outperforms the state-of-the-art methods at multiple noise levels.
   \blfootnote{Code: \url{https://github.com/Xu-Jingyi/FedCorr}}
\end{abstract}

\section{Introduction}
\label{sec:intro}
Federated learning (FL) is a promising solution for large-scale collaborative learning, where clients jointly train a machine learning model, while still maintaining local data privacy~\cite{mcmahan2017communication,yang2019survey,li2020survey}. However, in real-world FL implementations over heterogeneous networks, there may be differences in the characteristics of different clients due to diverse annotators' skill, bias, and hardware reliability~\cite{chen2020focus, yang2020robust}. Client data is rarely IID and frequently imbalanced. Also, some clients would have clean data, while other clients may have data with label noise at different noise levels. Hence, the deployment of practical FL systems would face challenges brought by discrepancies in two aspects i): local data statistics~\cite{mcmahan2017communication, li2020prox, hsu2019measuring,chen2021dynamic}, and ii): local label quality~\cite{chen2020focus,yang2020robust}. Although recent works explored the discrepancy in local data statistics in FL, and learning with label noise in centralized learning (CL), there is at present no unified approach for tackling both challenges simultaneously in FL.

The first challenge has been explored in recent FL works, with a focus on performance with convergence guarantees \cite{li2019noniid,reddi2021adaptive}. However, these works have the common implicit assumption that the given labels of local data are completely correct, which is rarely the case in real-world datasets. 

The second challenge can be addressed by reweighting~\cite{fu2019attack,chen2020focus,wan2021robust} or discarding~\cite{xu2021reputation} those client updates that are most dissimilar. In these methods, the corresponding clients are primarily treated as malicious agents. However, dissimilar clients are not necessarily malicious and could have label noise in local data that would otherwise still be useful after label correction.
\blfootnote{This work is supported by the National Research Foundation, Singapore under its AI Singapore Program (AISG Award No: AISG-RP-2019-015), and under its NRFF Program (NRFFAI1-2019-0005). This work is also supported in part by the SUTD Growth Plan Grant for AI.} For FL systems, the requirement of data privacy poses an inherent challenge for any label correction scheme. \emph{How can clients identify their noisy labels to be corrected without needing other clients to reveal sensitive information?} For example, \cite{yang2020robust} proposes label correction for identified noisy clients with the guidance of extra data feature information exchanged between clients and server, which may lead to privacy concerns.

\begin{figure*}[htp!] 
  \centering
  \includegraphics[width=1.0\linewidth]
{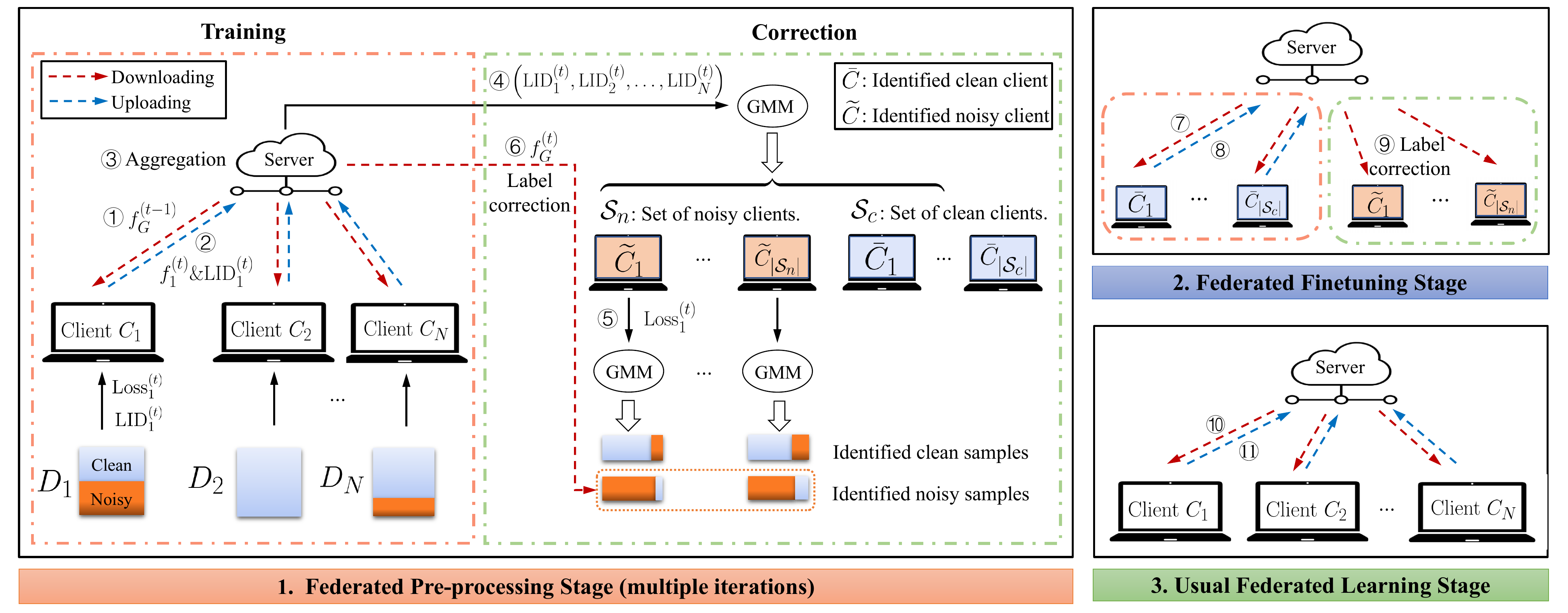}
   \caption{An overview of \texttt{FedCorr}, organized into three stages. Algorithm steps are numbered accordingly.}
   \label{fig:flow}
\end{figure*}

Label correction and, more generally, methods to deal with label noise, are well-studied in CL. Yet, even state-of-the-art CL methods for tackling label noise~\cite{Li2020DivideMix,arazo2019unsupervised,tanaka2018joint,han2020survey_reb,han2018co_reb,yu2019coteachingplus_reb,xia2020part_reb,xu2021training}, when applied to local clients, are inadequate in mitigating the performance degradation in the FL setting, due to the limited sizes of local datasets. These CL methods cannot be applied on the global sever or across multiple clients due to FL privacy requirements. So, it is necessary and natural to adopt a more general framework that jointly considers the two discrepancies, for a better emulation of real-world data heterogeneity. Most importantly, privacy-preserving label correction should be incorporated in training to improve robustness to data heterogeneity in FL. 

In this paper, we propose a multi-stage FL framework to simultaneously deal with both discrepancy challenges; see \cref{fig:flow} for an overview. To ensure privacy, we introduce a dimensionality-based filter to identify noisy clients, by measuring the local intrinsic dimensionality (LID) \cite{houle2013dimensionality} of local model prediction subspaces. Extensive experiments have shown that clean datasets can be distinguished from noisy datasets by the behavior of LID scores during training~\cite{ma2018dimensionality, ma2018characterizing}. Hence, in addition to the usual local weight updates, we propose that each client also sends an LID score to the server, which is a single scalar representing the discriminability of the predictions of the local model. We then filter noisy samples based on per-sample training losses independently for each identified noisy client, and relabel the large-loss samples with the predicted labels of the global model. To improve training stability and alleviate the negative impact caused by noisy clients, we introduce a weighted proximal regularization term, where the weights are based on the estimated local noise levels. Furthermore, we finetune the global model on the identified clean clients and relabel the local data for the remaining noisy clients. 

Our main contributions are as follows:
\begin{itemize}
    \item We propose a general multi-stage FL framework \texttt{FedCorr} to tackle data heterogeneity, with respect to both local label quality and local data statistics.
    \item We propose a general framework for easy generation of federated synthetic label noise and diverse (e.g. non-IID) client data partitions.
    \item We identify noisy clients via LID scores, and identify noisy labels via per-sample losses. We also propose an adaptive local proximal regularization term based on estimated local noise levels.
    \item We demonstrate that \texttt{FedCorr} outperforms state-of-the-art FL methods on multiple datasets with different noise levels, for both IID and non-IID data partitions. 
\end{itemize}

\section{Related work} 
\label{subsec:rela}
\subsection{Federated methods}
In this paper, we focus on three closely related aspects of FL: generation of non-IID federated datasets, methods to deal with non-IID local data, and methods for robust FL.

The generation of non-IID local data partitions for FL was first explored in \cite{mcmahan2017communication}, based on dividing a given dataset into shards. More recent non-IID data partitions are generated via Dirichlet distributions \cite{hsu2019measuring,acar2020federated,wan2021robust}. 

Recent federated optimization work mostly focus on dealing with the discrepancy in data statistics of local clients and related inconsistency issues ~\cite{li2020prox,wang2020tackling,acar2020federated}. For instance, \texttt{FedProx} deals with non-IID local data, by including a proximal term in the local loss functions~\cite{li2020prox}, while \texttt{FedDyn} uses a dynamic proximal term based on selected clients~\cite{acar2020federated}. \texttt{SCAFFOLD}~\cite{karimireddy2020scaffold} is another method suitable for non-IID local data that uses control variates to reduce client-drift. In \cite{hsu2019measuring} and \cite{reddi2021adaptive}, adaptive FL optimization methods for the global server are introduced, which are compatible with non-IID data distributions. Moreover, the Power-of-Choice (\texttt{PoC}) strategy~\cite{cho2020client}, a biased client selection scheme that selects clients with higher local losses, can be used to increase the rate of convergence.

There are numerous works on improving the robustness of FL; these include robust aggregation methods~\cite{wan2021robust,li2021ditto,fu2019attack}, reputation mechanism-based contribution examining~\cite{xu2021reputation}, credibility-based re-weighting~\cite{chen2020focus}, distillation-based semi-supervised learning~\cite{itahara2021distillation}, and personalized multi-task learning~\cite{li2021ditto}. However, these methods are not designed for identifying noisy labels. Even when these methods are used to detect noisy clients, either there is no mechanism for further label correction at the noisy clients~\cite{wan2021robust,li2021ditto,xu2021reputation,fu2019attack}, or the effect of noisy labels is mitigated with the aid of an auxiliary dataset, without any direct label correction~\cite{chen2020focus,itahara2021distillation}. One notable exception is \cite{yang2020robust}, which carries out label correction during training by exchanging feature centroids between clients and server. This exchange of centroids may lead to privacy concerns, since centroids could potentially be used as part of reverse engineering to reveal non-trivial information about raw local data.  

In contrast to these methods, \texttt{FedCorr} incorporates the generation of diverse local data distributions with synthetic label noise, together with noisy label identification and correction, without privacy leakage.

\subsection{Local intrinsic dimension (LID)}  
Informally, LID~\cite{houle2013dimensionality} is a measure of the intrinsic dimensionality of the data manifold. In comparison to other measures, LID has the potential for wider applications as it makes no further assumptions on the data distribution beyond continuity. The key underlying idea is that at each datapoint, the number of neighboring datapoints would grow with the radius of neighborhood, and the corresponding growth rate would then be a proxy for ``local'' dimension. 

LID builds upon this idea~\cite{houle2017local} via the geometric intuition that the volume of an $m$-dimensional Euclidean ball grows proportionally to $r^m$ when its radius is scaled by a factor of $r$. Specifically, when we have two $m$-dimensional Euclidean balls with volumes $V_1$, $V_2$, and with radii $r_1$, $r_2$, we can compute $m$ as follows:
\vspace*{-0.4em}
\begin{equation}
\label{eq: euclidean example}
    \frac{V_2}{V_1}=\bigg(\frac{r_2}{r_1}\bigg)^m \Rightarrow m=\frac{\log(V_2/V_1)}{\log(r_2/r_1)}.\\[{-0.3em}]
\end{equation}
We shall now formally define LID. Suppose we have a dataset consisting of vectors in $\mathbb{R}^n$. We shall treat this dataset as samples drawn from an $n$-variate distribution $\smash{\overline{\mathcal{D}}}$. For any $x \in \mathbb{R}^n$, let $Y_x$ be the random variable representing the (non-negative) distance from $x$ to a randomly selected point $y$ drawn from $\smash{\overline{\mathcal{D}}}$, and let $F_{Y_x}(t)$ be the cumulative distribution function of $Y_x$. Given $r>0$ and a sample point $x$ drawn from $\smash{\overline{\mathcal{D}}}$, define the \textit{LID of $x$ at distance $r$} to be 
\vspace*{-0.3em}
\[\LID_x(r) := \lim_{\varepsilon \to 0} \frac{\log F_{Y_x}((1+\varepsilon)r) - \log F_{Y_x}(r)}{\log (1+\varepsilon)},\\[{-0.3em}]\]
provided that it exists, i.e. provided that $F_{Y_x}(t)$ is positive and continuously differentiable at $t=r$. The \textit{LID at $x$} is defined to be the limit $\LID_x = \lim_{r\to 0} \LID_x(r)$. Intuitively, the LID at $x$ is an approximation of the dimension of a smooth manifold containing $x$ that would ``best'' fit the distribution $\smash{\overline{\mathcal{D}}}$ in the vincinity of $x$.

\textbf{Estimation of LID}: 
By treating the smallest neighbor distances as ``extreme events" associated to the lower tail of the underlying distance distribution, \cite{amsaleg2015estimating} proposes several estimators of LID based on extreme value theory. In particular, given a set of points $\mathcal{X}$, a reference point $x \in \mathcal{X}$, and its $k$ nearest neighbors in $\mathcal{X}$, the maximum-likelihood estimate (MLE) of $x$ is:
\vspace*{-0.4em}
\begin{equation}
    \label{eq: estimation of LID}
    \widehat{\text{LID}}(x)=-\bigg(\frac{1}{k}\smash{\sum_{i=1}^k}\log\frac{r_i(x)}{r_{max}(x)}\bigg)^{-1},
\end{equation}
\vspace*{-0.3em}
where $r_i(x)$ denotes the distance between $x$ and its $i$-th nearest neighbor, and $r_{max}(x)$ is the maximum distance from $x$ among the $k$ nearest neighbors.

\section{Proposed Method}
In this section, we introduce \texttt{FedCorr}, our proposed multi-stage training method to tackle heterogeneous label noise in FL systems (see Algorithm \ref{alg: fc}). Our method comprises three stages: pre-processing, finetuning and usual training. In the first stage, we sample the clients without replacement using a small fraction to identify noisy clients via LID scores and noisy samples via per-sample losses, after which we relabel the identified noisy samples with the predicted labels of the global model. The noise level of each client is also estimated in this stage. In the second stage, we finetune the model with a typical fraction on relatively clean clients, and use the finetuned model to further correct the samples for the remaining clients. Finally, in the last stage, we train the model via the usual FL method (\texttt{FedAvg}~\cite{mcmahan2017communication}) using the corrected labels at the end of the second stage. 

\begin{algorithm}[htb]
  \caption{\texttt{FedCorr} ({\color{red}Red} and Black line numbers in the pre-processing stage refer to operations for clients and server, respectively.) }
  \label{alg: fc}
  \textbf{Inputs:} $N$ (number of clients), $T_1$, $T_2$, $T_3$, $\mathcal{D}=\{ \mathcal{D}_i\}_{i=1}^N$ (dataset), \rlap{$w^{(0)}$ (initialized global model weights).} \\ 
  \textbf{Output:} Global model $f_G^{\text{final}}$ \\[-0.9em]
  \begin{algorithmic}[1] 
\STATEx{\textit{// Federated Pre-processing Stage} }
\STATE {$(\hat{\mu}_1^{(0)},\dots,\hat{\mu}_N^{(0)})\gets (0,\dots,0)$ \hfill // \textit{estimated noise levels}}
    \FOR {$t=1$ \textbf{to} $T_1$}
    \State{$\mathcal{S}= $Shuffle$(\{1,\dots,N\})$}
    \STATE{$w_{inter}\gets w^{(t-1)}$ \ \ \hfill // \textit{intermediary weights}}
    \FOR{$k\in\mathcal{S}$}
    \StateRed{$w_k^{(t)} \gets$ weights that minimize loss function \eqref{eq: loss}}
     \StateRed{Upload weights $w_k^{(t)}$ and LID score to server}
    \ENDFOR
    \StateBlack{Update global model $w^{(t)}\gets w_{inter}$}
    \STATE {Divide all clients into clean set $\mathcal{S}_c$ and noisy set $\mathcal{S}_n$}
    {based on cumulative LID scores via GMM}
    \FOR{\textbf{noisy client} $k \in \mathcal{S}_n$}
     \StateRed{Divide $\mathcal{D}_k$ into clean subset $\mathcal{D}^c_k$ and noisy subset $\mathcal{D}^n_k$ based on per-sample losses via GMM}
     \StateRed{ $\hat{\mu}_k^{(t)}\gets\frac{\vert \mathcal{D}^n_k\vert}{\vert \mathcal{D}_k \vert}$ \hfill \textit{// update estimated noise level}}
     \StateRed{ $y_k^{(i)}\gets \argmax f(x_k^{(i)};w^{(i)})$,   $\forall (x_k^{(i)},y_k^{(i)})\in \mathcal{D}_k^n$}
    \ENDFOR
\ENDFOR

\STATEx{\vspace*{-0.9em}}
\STATEx{\textit{// Federated Finetuning Stage}}
\StateBlack{$\mathcal{S}_c\gets \{k\vert k\in\mathcal{S}, \mu_k<0.1\}$,  $\mathcal{S}_n\gets\mathcal{S}\setminus\mathcal{S}_c$.}
    \FOR {$t=T_1+1$ \textbf{to} $T_1+T_2$}
  \STATE{Update $w_k^{(t)}$ by usual \texttt{FedAvg} among clients in $\mathcal{S}_c$}
    \ENDFOR
\FOR{\textbf{Noisy client} $k \in \mathcal{S}_n$}
    \STATE{ $y_k^{(i)}\gets \argmax f(x_k^{(i)};w^{(i)})$, $\forall (x_k^{(i)},y_k^{(i)})\in \mathcal{D}_k$}
\ENDFOR    

\STATEx{\vspace*{-0.9em}}
\STATEx{\textit{// Usual Federated Learning Stage}}
    \FOR {$t=T_1+T_2+1$ \textbf{to} $T_1+T_2+T_3$}
    \STATE{Update $w_k^{(t)}$ by usual \texttt{FedAvg} among all clients}
    \ENDFOR
\RETURN {$f_G^{\text{final}}:=f(\cdot;w^{(T_1+T_2+T_3)})$}
\end{algorithmic}

\end{algorithm}     
\subsection{Preliminaries}
\label{notation and problem stagements}
Consider an FL system with $N$ clients and an $M$-class dataset $\mathcal{D}=\{\mathcal{D}_k\}_{k=1}^N$, where each $\mathcal{D}_k=\{(x_k^i, y_k^i)\}_{i=1}^{n_k}$ denotes the local dataset for client $k$. Let $\mathcal{S}$ denote the set of all $N$ clients, and let $w^{(t)}_k$ (resp. $w^{(t)}$) denote the local model weights of client $k$ (resp. global model weights obtained by aggregation) at the end of communication round $t$. At the end of round $t$, the global model $f_G^{(t)}$ would have its weights $w^{(t)}$ updated as follows: 
\vspace*{-0.3em}
\begin{equation}
    \label{eq: aggregation}
    w^{(t)} \leftarrow \sum_{\smash{k \in \mathcal{S}_t}} \frac{\smash{|\mathcal{D}_k|}}{\sum_{i \in \mathcal{S}_t}|\mathcal{D}_i|} w_{k}^{(t)}, 
\end{equation}
\vspace*{-0.2em}
where $\mathcal{S}_t \subseteq \mathcal{S}$ is the subset of selected clients in round $t$.

For the rest of this subsection, we shall give details on client data partition, noise model simulation, and LID score computation.
These are three major aspects of our proposed approach to emulate data heterogeneity, and to deal with the discrepancies in both local data statistics and label quality.

\begin{figure}[htb]
  \centering
  \includegraphics[width=1.0\columnwidth]
{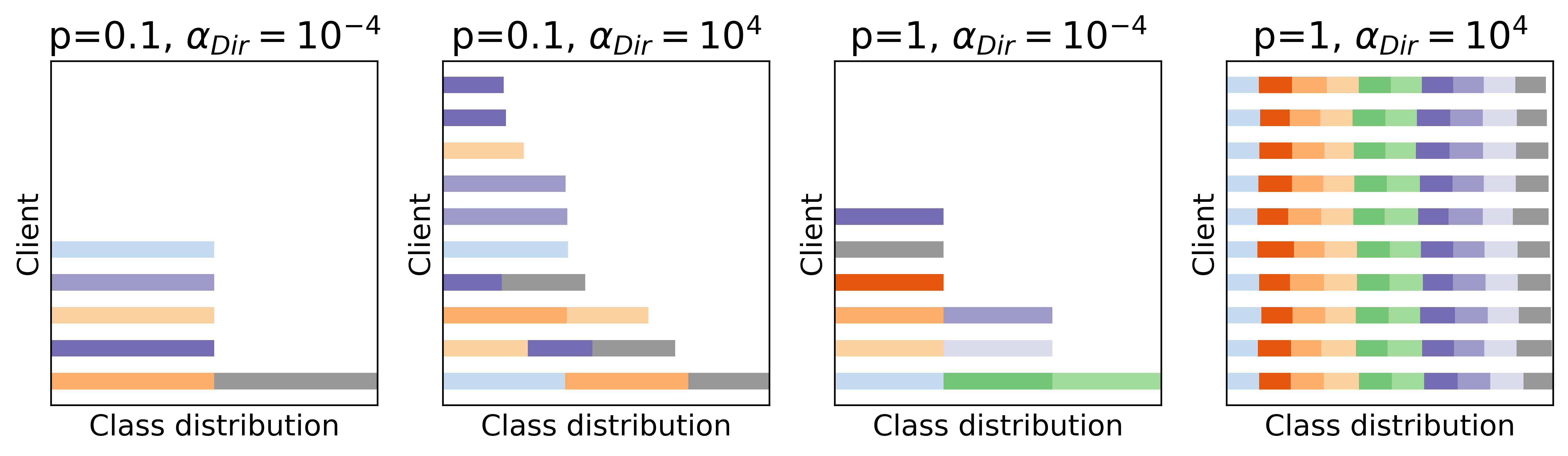}

   \caption{Depiction of non-IID partitions for different parameters.}
   \label{fig:noniid}
\end{figure}

\textbf{Data partition}.
We consider both IID and non-IID heterogeneous data partitions in this work. For IID partitions, the whole dataset $\mathcal{D}$ is uniformly distributed at random among $N$ clients. 
For non-IID partitions, we first generate an $N\times M$ indicator matrix $\Phi$, where each entry $\Phi_{ij}$ indicates whether the local dataset of client $i$ contains class $j$. 
Each $\Phi_{ij}$ shall be sampled from the Bernoulli distribution with a fixed probability $p$. 
For each $1\leq j\leq M$, let $\upsilon_j$ be the sum of entries in the $j$-th column of $\Phi$; this equals the number of clients whose local datasets contain class $j$. Let \smash{$\boldsymbol{q}_j$} be a vector of length $\upsilon_j$, sampled from the symmetric Dirichlet distribution with the common parameter $\alpha_{Dir}>0$. Using \smash{$\boldsymbol{q}_j$} as a probability vector, we then randomly allocate the samples within class $j$ to these $\upsilon_j$ clients. Note that our non-IID data partition method provides a general framework to control the variability in both class distribution and the sizes of local datasets (see \cref{fig:noniid}).

\vspace*{-1.3em}
\textbf{Noise model}.
To emulate label noise in real-world data, we shall introduce a general federated noise model framework. 
For simplicity, this work only considers instance-independent label noise. This framework has two parameters $\rho$ and $\tau$, where $\rho$ denotes the system noise level (ratio of noisy clients) and $\tau$ denotes the lower bound for the noise level of a noisy client. Every client has a probability $\rho$ of being a noisy client, in which case the local noise level for this noisy client is determined randomly, by sampling from the uniform distribution $\textit{U}(\tau,1)$. Succinctly, the noise level of client $k$ (for $ k=1,\dots,N$) is
\vspace*{-0.3em}
\begin{equation}
\label{mu}
    \mu_k=
    \begin{cases}
     u\sim\textit{U}(\tau,1),&\text{with probability }\rho; \\
    0, &\text{with probability }1-\rho.
    \end{cases}
\end{equation}
When $\mu_k \neq 0$, the $100\cdot\mu_k\%$ noisy samples are chosen uniformly at random, and are assigned random labels, selected uniformly from the $M$ classes. 

\textbf{LID scores for local models}.
In this paper, we associate LID scores to local models. Consider an arbitrary client with local dataset $\mathcal{D}$ and current local model $f(\cdot)$. Let $\mathcal{X} := \{f(x)\}_{\smash{x\in \mathcal{D}}}$ be the set of prediction vectors, and for each $x\in \mathcal{D}$, compute $\widehat{LID}(f(x))$ w.r.t. the $k$ nearest neighbors in $\mathcal{X}$, as given in \eqref{eq: estimation of LID}. We define the \textit{LID score} of $(\mathcal{D},f)$ to be the average value of $\widehat{LID}(f(x))$ over all $x\in \mathcal{D}$. Note that as the local model $f(\cdot)$ gets updated with each round, the corresponding LID score will change accordingly.

Experiments have shown that given the same training process, models trained on a dataset with label noise tend to have larger LID scores as compared to models trained on the same dataset with clean labels~\cite{ma2018dimensionality, ma2018characterizing}.  Intuitively, the prediction vectors of a well-trained model, trained on a clean dataset, would cluster around $M$ possible one-hot vectors, corresponding to the $M$ classes. However, as more label noise is added to the clean dataset, the prediction vector of a noisy sample would tend to be shifted towards the other clusters, with different noisy samples shifted in different directions. Hence, the prediction vectors near each one-hot vector would become ``more diffuse'' and would on average span a higher dimensional space.

\subsection{Federated pre-processing stage}
\texttt{FedCorr} begins with the pre-processing stage, which iteratively evaluates the quality of the dataset of each client, and relabels identified noisy samples. This pre-processing stage differs from traditional FL in the following aspects: 
\begin{itemize}
    \item  All clients will participate in each iteration. Clients are selected without replacement, using a small fraction.
    \item  An adaptive local proximal term is added to the loss function, and mixup data augmentation is used.
    \item  Each client computes its LID score and per-sample cross-entropy loss after local training and sends its LID score together with local model updates to the server.
\end{itemize}

\vspace*{0.4em}
\noindent\textbf{Client iteration and fraction scheduling.} \\
The pre-processing stage is divided into $T_1$ iterations. In each iteration, every client participates exactly once. Every iteration is organized by communication rounds, similar to the usual FL, but with two key differences: a small fraction is used, and clients are selected without replacement. Each iteration ends when all clients have participated.

It is known that large fractions could help improve the convergence rate\cite{mcmahan2017communication}, and a linear speedup could even be achieved in the case of convex loss functions\cite{stich2018local}. 
However, large fractions have a weak effect in non-IID settings, while intuitively, small fractions would yield aggregated models that deviate less from local models; cf. \cite{li2019convergence}. These observations inspire us to propose a fraction scheduling scheme that combines the advantages of both small and large fractions. Specifically, we sample clients using a small fraction without replacement in the pre-processing stage, and use a typical larger fraction with replacement in the latter two stages. By sampling without replacement during pre-processing, we ensure all clients participate equally for the evaluation of the overall quality of labels in local datasets.

\vspace*{0.4em}
\noindent\textbf{Mixup and local proximal regularization.} \\
Throughout the pre-processing stage, for client $k$ with batch $(X_b, Y_b)=\{(x_i, x_j)\}_{i=1}^{n_b}$ (where $n_b$ denotes batch size), we use the following loss function:
\vspace*{-0.3em}
\begin{equation}
\label{eq: loss}
\small
    L(X_b) = L_{CE}\left(f_k^{(t)}(\tilde{X}_b), \tilde{Y}_b\right) + \beta\hat{\mu}_k^{(t-1)}\big\| w_k^{(t)}-w^{(t-1)} \big\|^2. \\[{-0.1em}]
\end{equation} 
Here, \smash{$f_k^{(t)}=f(\cdot;w_k^{(t)})$} denotes the local model of client $k$ in round $t$, and $w^{(t-1)}$ denotes the weights of the global model obtained in the previous round $t-1$. The first term in \eqref{eq: loss} represents the cross-entropy loss on the mixup augmentation of $(X_b, Y_b)$, while the second term in \eqref{eq: loss} is an adaptive local proximal regularization term, where $\hat{\mu}_k^{(t-1)}$ is the estimated noise level of client $k$ to be defined later. It should be noted that our local proximal regularization term is only applied in the pre-processing stage. 

Recall that mixup \cite{zhang2018mixup} is a data augmentation technique that favors linear relations between samples, and that has been shown to exhibit strong robustness to label noise \cite{arazo2019unsupervised, Li2020DivideMix}. Mixup generates new samples $(\tilde{x}, \tilde{y})$ as convex combinations of randomly selected pairs of samples $(x_i, y_i)$ and $(x_j, y_j)$, given by $\tilde{x}=\lambda x_i + (1-\lambda) x_j$, $\tilde{y}=\lambda y_i + (1-\lambda) y_j$, where $\lambda\sim \text{Beta}(\alpha, \alpha)$, and $\alpha\in (0, \infty)$. (We use $\alpha=1$ in our experiments.) Intuitively, mixup achieves robustness to label noise due to random interpolation. For example, if $(x_i, \hat{y}_i)$ is a noisy sample and if $y_i$ is the true label, then the negative impact caused by an incorrect label $\hat{y}_i$ is alleviated when paired with a sample whose label is $y_i$.

Our adaptive local proximal regularization term is scaled by \smash{$\hat{\mu}_k^{(t-1)}$}, which is the estimated noise level of client $k$ computed at the end of round $t-1$. (In particular, this term would vanish for clean clients.)  The hyperparameter $\beta$ is also incorporated to control the overall effect of this term. Intuitively, if a client's dataset has a larger discrepancy from other local datasets, then the corresponding local model would deviate more from the global model, thereby contributing a larger loss value for the local proximal term. 

\vspace*{0.4em}
\noindent\textbf{Identification of noisy clients and noisy samples.} \\
To address the challenge of heterogeneous label noise, we shall iteratively identify and relabel the noisy samples. In each iteration of this pre-processing stage, where all clients will participate, every client will compute the LID score and per-sample loss for its current local model (see \cref{alg: fc}, lines 3-9). Specifically, when client $k$ is selected in round $t$, we train the model \smash{$f_k^{(t)}$} on the local dataset \smash{$\mathcal{D}_k$} and then compute the LID score of \smash{$(\mathcal{D}_k,f_k^{(t)})$} via \eqref{eq: estimation of LID}. Note that our proposed framework preserves the privacy of client data, since in comparison to the usual FL, there is only an additional LID score sent to the server, which is a single scalar that reflects only the predictive discriminability of the local model. Since the LID score is computed from the predictions of the output layer (of the local model), knowing this LID score does not reveal information about the raw input data. This additional LID score is a single scalar, hence it has a negligible effect on communication cost.

At the end of iteration $t$, we shall perform the following three steps:
\begin{enumerate}
    \item The server first computes a Gaussian Mixture Model (GMM) on the cumulative LID scores of all $N$ clients. Using this GMM, the set of clients $\mathcal{S}$ is partitioned into two subsets: \smash{$\mathcal{S}_n$} (noisy clients) and \smash{$\mathcal{S}_c$} (clean clients). 
    \item Each noisy client \smash{$k\in\mathcal{S}_n$} locally computes a new GMM on the per-sample loss values for all samples in the local dataset \smash{$\mathcal{D}_k$}. Using this GMM, \smash{$\mathcal{D}_k$} is partitioned into two subsets: a clean subset \smash{$\mathcal{D}_k^c$}, and a noisy subset \smash{$\mathcal{D}_k^n$}. We observe that the large-loss samples are more likely to have noisy labels. 
    The local noise level of client $k$ can then be estimated by \smash{$\hat{\mu}_k^{(t)} = |\mathcal{D}_k^n|/|\mathcal{D}_k|$} if \smash{$k\in \mathcal{S}_n$}, and \smash{$\hat{\mu}_k^{(t)} = 0$} otherwise.
    \item Each noisy client \smash{$k \in \mathcal{S}_n$} performs relabeling of the noisy samples by using the predicted labels of the global model as the new labels. 
    In order to avoid over-correction, we only relabel those samples that are identified to be noisy with high confidence. This partial relabeling is controlled by a relabel ratio $\pi$ and a confidence threshold $\theta$. 
    Take noisy client $k$ for example: We first choose samples from \smash{$\mathcal{D}_k^n$} that corresponds to the top-$\pi\cdot|\mathcal{D}_k^n|$ largest per-sample cross-entropy losses. Next, we obtain the prediction vectors of the global model, and relabel a sample only when the maximum entry of its prediction vector exceeds $\theta$. Thus, the subset \smash{$\widetilde{\mathcal{D}_k^n}'$} of samples to be relabeled is given by
    \vspace*{-0.4em}
    \begin{align}
  \smash{\widetilde{\mathcal{D}_k^n}} &= \argmax_{\tilde{\mathcal{D}}\subseteq\mathcal{D}_k^n\atop\vert\tilde{\mathcal{D}}\vert=\pi\cdot\vert\mathcal{D}_k^n\vert} L_{CE}(\tilde{\mathcal{D}}; f_G^{(t)});  \\
  \label{eq: confidence threshold}
  \smash{\widetilde{\mathcal{D}_k^n}'} &= \big\{(x,y)\in \smash{\widetilde{\mathcal{D}_k^n}}\big | \max(f_G^{(t)}(x))\geq\theta \big\};\\[-1.8em]\nonumber {}
    \end{align}
    where $f_G^{(t)}$ is the global model at the end of iteration $t$.
\end{enumerate}
\vspace*{0.1em}
\noindent\textit{Why do we use cumulative LID scores in step 1?} \\
In deep learning, it has been empirically shown that when training on a dataset with label noise, the evolution of the representation space of the model exhibits two distinct phases: (1) an early phase of dimensionality compression, where the model tends to learn the underlying true data distribution, and (2) a later phase of dimensionality expansion, where the model overfits to noisy labels \cite{ma2018dimensionality}. 

\begin{figure}[t!]
  \centering
  \includegraphics[width=1\columnwidth]
{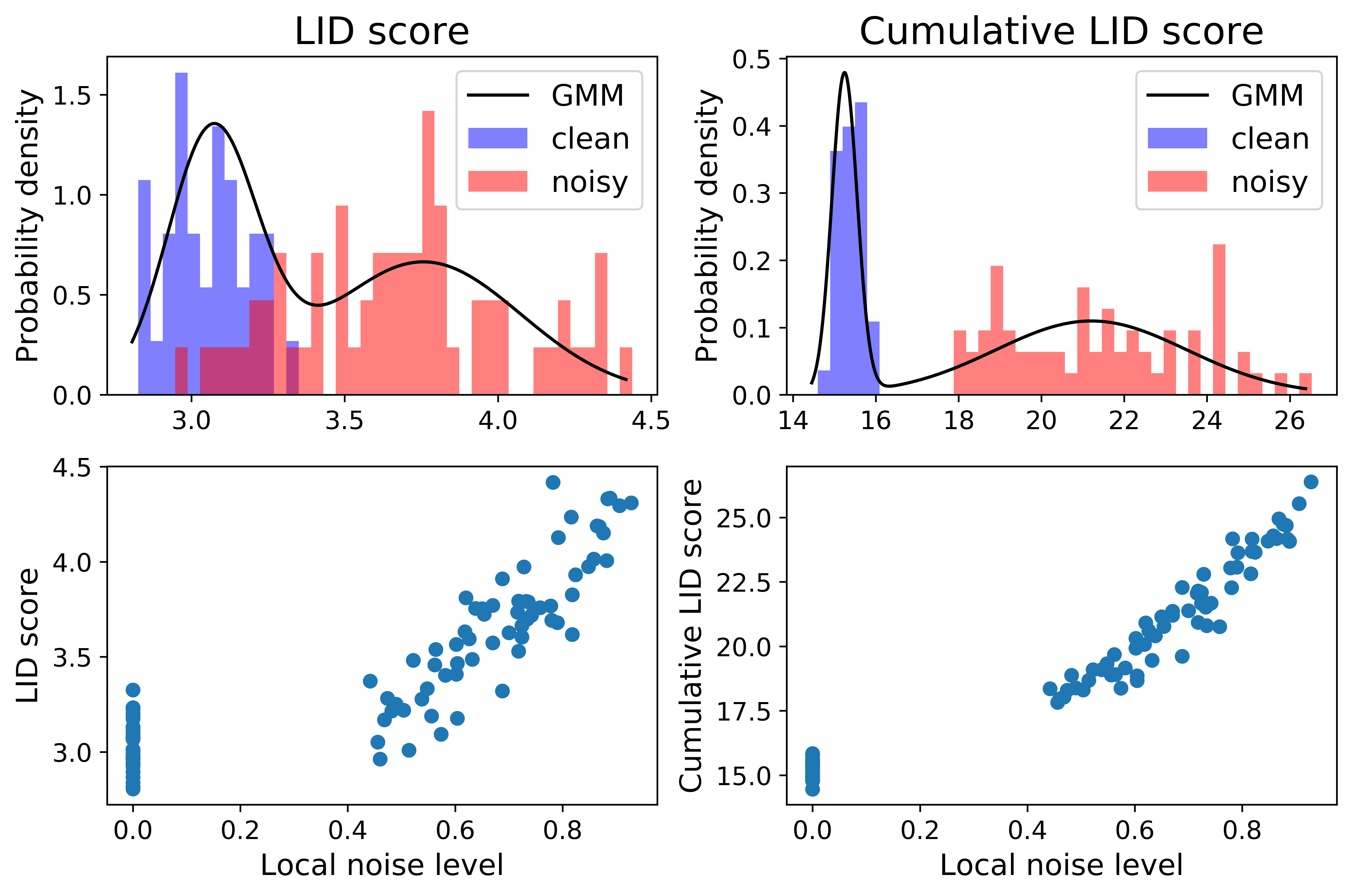}
   \caption{Empirical evaluation of LID score (left) and cumulative LID score (right) after 5 iterations on CIFAR-10 with noise model $(\rho,\tau)=(0.6,0.5)$, and with IID data partition, over 100 clients. Top: probability density function and estimated GMM; bottom: LID/cumulative LID score vs. local noise level for each client.}
   \label{fig: lid}
\end{figure}

We observed that clients with larger noise levels tend to have larger LID scores. Also, the overlap of LID scores between clean and noisy clients would increase during training. This increase could be due to two reasons: (1) the model may gradually overfit to noisy labels, and (2) we correct the identified noisy samples after each iteration, thereby making the clients with low noise levels less distinguishable from clean clients. Hence, the cumulative LID score (i.e., the sum of LID scores in all past iterations) is a better metric for distinguishing noisy clients from clean clients; see the top two plots in \cref{fig: lid} for a comparison of using LID score versus cumulative LID score. Furthermore, the bottom two plots in \cref{fig: lid} show that cumulative LID score has a stronger linear relation with local noise level.

\subsection{Federated finetuning stage}
We aim to finetune the global model $f_G$ on relatively clean clients over $T_2$ rounds and further relabel the remaining noisy clients. The aggregation at the end of round $t$ is given by the same equation \eqref{eq: aggregation}, with one key difference: $\mathcal{S}_t$ is now a subset of $\mathcal{S}_c=\{k|1\leq k\leq N, \hat{\mu}_k^{(T_1)}\leq\kappa\}$, where $\kappa$ is the threshold used to select relatively clean clients based on the estimated local noise levels $\hat{\mu}_1^{(T_1)},...,\hat{\mu}_N^{(T_1)}$.

At the end of the finetuning stage, we relabel the remaining noisy clients $\mathcal{S}_n=\mathcal{S}\setminus\mathcal{S}_c$ with the predicted labels of $f_G$. Similar to the correction process in the pre-processing stage, we use the same confidence threshold $\theta$ to control the subset of samples to be relabeled; see \eqref{eq: confidence threshold}. 

\subsection{Federated usual training stage}
In this final stage, we train the global model over $T_3$ rounds via the usual FL (\texttt{FedAvg}) on all the clients, using the labels corrected in the previous two training stages. We also incorporate this usual training stage with three FL methods to show that methods based on different techniques can be well-incorporated with \texttt{FedCorr}, even if they are not designed specifically for robust FL; see \cref{subsec: exp}.

\begin{table}[t!]
  \centering
  \begin{adjustbox}{width=\columnwidth,center}
  \begin{tabular}{lccc}
    \toprule
    \midrule
    Dataset & CIFAR-10 & CIFAR-100 & Clothing1M \\
    \midrule
    Size of $\mathcal{D}_{train}$ & 50,000 & 50,000 & 1,000,000 \\
    \# of classes & 10 & 100 & 14 \\
    \# of clients & 100 & 50 & 500 \\
    Fraction $\gamma$ & 0.1 & 0.1 & 0.02\\
    Architecture & ResNet-18 & ResNet-34 & pre-trained ResNet-50\\
    \bottomrule
  \end{tabular}
  \end{adjustbox}
  \caption{List of datasets used in our experiments.}
  \label{tab:dataset}
\end{table}

\section{Experiments}
\label{sec: exp}

\begin{table*}[!ht]
  \centering
  \begin{adjustbox}{width=\textwidth,center}
  \begin{tabular}{ll|rrrrrrr}
    \toprule
    \midrule
     \multirow{3}{*}{Setting} & \multirow{3}{*}{Method} & \multicolumn{7}{|c}{Best Test Accuracy (\%) $\pm$ Standard Deviation (\%)} \\
    \cmidrule{3-9}
     & & \multicolumn{1}{|c|}{$\rho=0.0$} &\multicolumn{2}{|c|}{$\rho=0.4$} & \multicolumn{2}{c}{$\rho=0.6$} & \multicolumn{2}{|c}{$\rho=0.8$} \\ \cmidrule{3-9} 
     & & \multicolumn{1}{c|}{$\tau=0.0$} & \multicolumn{1}{c|}{$\tau=0.0$} & \multicolumn{1}{c|}{$\tau=0.5$} & \multicolumn{1}{c|}{$\tau=0.0$} & \multicolumn{1}{c|}{$\tau=0.5$} & \multicolumn{1}{c|}{$\tau=0.0$} & \multicolumn{1}{c}{$\tau=0.5$} \\
    \midrule
    \multirow{2}{*}{\makecell[l]{Centralized \\ (for reference)}} & JointOpt  &  93.73$\pm$0.21& 92.29$\pm$0.37& 92.11$\pm$0.21& 91.26$\pm$0.46& 88.42$\pm$0.33& 89.18$\pm$0.29& 85.62$\pm$1.17 \\
    & DivideMix &  95.64$\pm$0.05& 96.39$\pm$0.09& 96.17$\pm$0.05& 96.07$\pm$0.06& 94.59$\pm$0.09& 94.21$\pm$0.27& 94.36$\pm$0.16 \\
    \midrule
      \multirow{6}{*}{\makecell[l]{Federated }} & FedAvg & 93.11$\pm$0.12& 89.46$\pm$0.39& 88.31$\pm$0.80& 86.09$\pm$0.50& 81.22$\pm$1.72& 82.91$\pm$1.35& 72.00$\pm$2.76 \\
      & FedProx  &  92.28$\pm$0.14& 88.54$\pm$0.33& 88.20$\pm$0.63& 85.80$\pm$0.41& 85.25$\pm$1.02& 84.17$\pm$0.77& 80.59$\pm$1.49 \\
      & RoFL    & 88.33$\pm$0.07 & 88.25$\pm$0.33 & 87.20$\pm$0.26 & 87.77$\pm$0.83 & 83.40$\pm$1.20 & 87.08$\pm$0.65 & 74.13$\pm$3.90  \\
      & ARFL  &92.76$\pm$0.08  &85.87$\pm$1.85  &83.14$\pm$3.45  &76.77$\pm$1.90  &64.31$\pm$3.73  &73.22$\pm$1.48 & 53.23$\pm$1.67 \\
      & JointOpt  &  88.16$\pm$0.18& 84.42$\pm$0.70& 83.01$\pm$0.88& 80.82$\pm$1.19& 74.09$\pm$1.43& 76.13$\pm$1.15& 66.16$\pm$1.71 \\
      & DivideMix &77.96$\pm$0.15&77.35$\pm$0.20&74.40$\pm$2.69&72.67$\pm$3.39&72.83$\pm$0.30&68.66$\pm$0.51&68.04$\pm$1.38\\
      &  Ours      & \textbf{93.82$\pm$0.41}& \textbf{94.01$\pm$0.22}& \textbf{94.15$\pm$0.18}& \textbf{92.93$\pm$0.25}& \textbf{92.50$\pm$0.28}& \textbf{91.52$\pm$0.50}& \textbf{90.59$\pm$0.70}   \\ 
    \bottomrule
  \end{tabular}
  \end{adjustbox}
  \caption{Average (5 trials) and standard deviation of the best test accuracies of various methods on CIFAR-10 with IID setting at different noise levels ($\rho$: ratio of noisy clients, $\tau$: lower bound of client noise level). The highest accuracy for each noise level is boldfaced.}
  \label{experiment: cifar-10 IID}
\end{table*}

\begin{table}[ht!]
  \centering
  \begin{adjustbox}{width=\columnwidth,center}
  \begin{tabular}{l|rrrr}
    \toprule
    \midrule
     \multirow{3}{*}{Method} & \multicolumn{4}{|c}{Best Test Accuracy (\%) $\pm$ Standard Deviation(\%)} \\
    \cmidrule{2-5}
    &\multicolumn{1}{c|}{$\rho=0.0$}  &\multicolumn{1}{c|}{$\rho=0.4$} &\multicolumn{1}{c|}{$\rho=0.6$}  &\multicolumn{1}{c}{$\rho=0.8$}\\
    &\multicolumn{1}{c|}{$\tau=0.0$}  &\multicolumn{1}{c|}{$\tau=0.5$} &\multicolumn{1}{c|}{$\tau=0.5$} &\multicolumn{1}{c}{$\tau=0.5$}\\
    \midrule
      JointOpt (CL)&72.94$\pm$0.43&65.87$\pm$1.50 &60.55$\pm$0.64&59.79$\pm$2.45 \\
      DivideMix (CL)&75.58$\pm$0.14&75.43$\pm$0.34&72.26$\pm$0.58&71.02$\pm$0.65\\
    \midrule
      FedAvg & 72.41$\pm$0.18& 64.41$\pm$1.79& 53.51$\pm$2.85& 44.45$\pm$2.86 \\
      FedProx  &  71.93$\pm$0.13& 65.09$\pm$1.46& 57.51$\pm$2.01& 51.24$\pm$1.60 \\
      RoFL    &  67.89$\pm$0.65& 59.42$\pm$2.69& 46.24$\pm$3.59& 36.65$\pm$3.36 \\
      ARFL  & 72.05$\pm$0.28 &51.53$\pm$4.38  &33.03$\pm$1.81  &27.47$\pm$1.08  \\
      JointOpt  &  67.49$\pm$0.36& 58.43$\pm$1.88& 44.54$\pm$2.87& 35.25$\pm$3.02  \\
      DivideMix &45.91$\pm$0.27 &43.25$\pm$1.01 &40.72$\pm$1.41 &38.91$\pm$1.25  \\
      Ours   &\textbf{72.56$\pm$2.07}&\textbf{74.43$\pm$0.72}  &\textbf{66.78$\pm$4.65}  &\textbf{59.10$\pm$5.12} \\ 
    \bottomrule
  \end{tabular}
  \end{adjustbox}
  \caption{Average (5 trials) and standard deviation of the best test accuracies on CIFAR-100 with IID setting.}
  \label{experiment: cifar-100}
\end{table}

\begin{table}[ht!]
  \centering
  \begin{adjustbox}{width=\columnwidth,center}
  \begin{tabular}{lccc}
    \toprule
    \midrule
    Method$\backslash$($p,\alpha_{Dir}$) & $(0.7,10)$ &$(0.7,1)$&$(0.3,10)$\\
    \midrule
    FedAvg &78.88$\pm$2.34&75.98$\pm$2.92&67.75$\pm$4.38\\
    FedProx &83.32$\pm$0.98&80.40$\pm$0.94&73.86$\pm$2.41  \\
    RoFL  & 79.56$\pm$1.39& 72.75$\pm$2.21& 60.72$\pm$3.23 \\
    ARFL  &60.19$\pm$3.33&55.86$\pm$3.30&45.78$\pm$2.84\\
    JointOpt &72.19$\pm$1.59&66.92$\pm$1.89&58.08$\pm$2.18  \\
    DivideMix &65.70$\pm$0.35&61.68$\pm$0.56&56.67$\pm$1.73 \\
    Ours &\textbf{90.52$\pm$0.89}&\textbf{88.03$\pm$1.08}&\textbf{81.57$\pm$3.68}  \\
    \bottomrule
  \end{tabular}
  \end{adjustbox}
  \caption{Average (5 trials) and standard deviation of the best test accuracies of different methods on CIFAR-10 with different non-IID setting. The noise level is $(\rho,\tau)=(0.6,0.5)$. }
  \label{tab:cifar10_noniid}
\end{table}

\begin{table}[ht!]
  \centering
  \begin{adjustbox}{width=\columnwidth,center}
  \begin{tabular}{lccccccc}
    \toprule
    \midrule
    Settings  & FedAvg &FedProx &RoFL &ARFL &JointOpt &Dividemix &Ours \\
    \midrule
    FL  & 70.49 & 71.35 & 70.39 & 70.91 & 71.78 & 68.83 & \textbf{72.55} \\
    CL   & - & - & - & - & 72.23 & 74.76 & - \\
    \bottomrule
  \end{tabular}
  \end{adjustbox}
  \caption{Best test accuracies on Clothing1M with non-IID setting. CL results are the accuracies reported in corresponding papers.}
  \label{table: clothing_noniid}
\end{table}

In this section, we conduct experiments in both IID (CIFAR-10/100 \cite{krizhevsky2009learning}) and non-IID (CIFAR-10, Clothing1M\cite{xiao2015learning}) data settings, at multiple noise levels, to show that \texttt{FedCorr} is simultaneously robust to both local label quality discrepancy and data statistics discrepancy. To demonstrate the versatility of \texttt{FedCorr}, we also show that various FL methods can have their performances further improved by incorporating the first two stages of \texttt{FedCorr}. We also conduct an ablation study to show the effects of different components of \texttt{FedCorr}. Details on data partition and the noise model used have already been given in \cref{notation and problem stagements}.

\subsection{Experimental Setup}
\label{subsec: expset}
\noindent\textbf{Baselines.} There are two groups of experiments. 

In the first group, we demonstrate that \texttt{FedCorr} is robust to discrepancies in both data statistics and label quality. We compare \texttt{FedCorr} with the following state-of-the-art methods from three categories: (1) methods to tackle label noise in CL (\texttt{JointOpt} \cite{tanaka2018joint} and \texttt{DivideMix} \cite{Li2020DivideMix}) applied to local clients; (2) classic FL methods (\texttt{FedAvg}\cite{mcmahan2017communication} and \texttt{FedProx}\cite{li2020prox}); and (3) FL methods designed to be robust to label noise (\texttt{RoFL}\cite{yang2020robust} and \texttt{ARFL}\cite{fu2019attack}). 
For reference, we also report experimental results on \texttt{JointOpt} and \texttt{DivideMix} in CL, so as to show the performance reduction of these two methods when used in FL.

In the second group, we demonstrate the versatility of \texttt{FedCorr}. We examine the performance improvements of three state-of-the-art methods when the first two stages of \texttt{FedCorr} are incorporated.
These methods are chosen from three different aspects to improve FL: local optimization (\texttt{FedDyn}~\cite{acar2020federated}), aggregation (\texttt{Median}~\cite{pmlr-v80-yin18a}) and client selection (\texttt{PoC}~\cite{cho2020client}).

\begin{table*}[ht!]
  \centering
  \begin{adjustbox}{width=\textwidth,center}
  \begin{tabular}{l|ccccccc}
    \toprule
    \midrule
    \multirow{3}{*}{Method} & \multicolumn{7}{|c}{Best Test Accuracy (\%) $\pm$ Standard Deviation (\%)} \\
    \cmidrule{2-8}
     & \multicolumn{1}{|c|}{$\rho=0.0$} & \multicolumn{2}{|c|}{$\rho=0.4$} & \multicolumn{2}{c}{$\rho=0.6$} & \multicolumn{2}{|c}{$\rho=0.8$} \\ \cmidrule{2-8} 
     & \multicolumn{1}{c|}{$\tau=0.0$} & \multicolumn{1}{c|}{$\tau=0.0$} & \multicolumn{1}{c|}{$\tau=0.5$} & \multicolumn{1}{c|}{$\tau=0.0$} & \multicolumn{1}{c|}{$\tau=0.5$} & \multicolumn{1}{c|}{$\tau=0.0$} & \multicolumn{1}{|c}{$\tau=0.5$} \\
    \midrule
      \multicolumn{1}{l|}{Ours}  & \textbf{93.82$\pm$0.41}& \textbf{94.01$\pm$0.22}& \textbf{94.15$\pm$0.18}& \textbf{92.93$\pm$0.25}& \textbf{92.50$\pm$0.28}& \textbf{91.52$\pm$0.50}& \textbf{90.59$\pm$0.70} \\
      \midrule
      \multicolumn{1}{l|}{Ours w/o correction}  &  92.85$\pm$0.66& 93.71$\pm$0.20& 93.60$\pm$0.21& 92.15$\pm$0.29& 91.77$\pm$0.65& 90.48$\pm$0.56& 88.77$\pm$1.10  \\
      \multicolumn{1}{l|}{Ours w/o frac. scheduling}     &  86.05$\pm$1.47& 85.59$\pm$1.10& 78.44$\pm$7.90& 80.29$\pm$2.62& 77.96$\pm$3.65& 76.67$\pm$3.48& 72.71$\pm$5.03 \\
      \multicolumn{1}{l|}{Ours w/o local proximal}    &  93.37$\pm$0.05& 93.64$\pm$0.15& 93.46$\pm$0.17& 92.34$\pm$0.14& 91.74$\pm$0.47& 90.45$\pm$0.94& 88.74$\pm$1.72 \\
      \multicolumn{1}{l|}{Ours w/o finetuning} & 92.71$\pm$0.18& 93.06$\pm$0.15& 92.62$\pm$0.28& 91.41$\pm$0.14& 89.31$\pm$0.90& 89.62$\pm$0.40& 83.81$\pm$2.59 \\
      \multicolumn{1}{l|}{Ours w/o usual training}      &  93.11$\pm$0.10& 93.53$\pm$0.17& 93.46$\pm$0.14& 92.16$\pm$0.24& 91.50$\pm$0.51& 90.62$\pm$0.59& 88.97$\pm$1.37 \\ 
      \multicolumn{1}{l|}{Ours w/o mixup}      &  90.63$\pm$0.70& 88.83$\pm$1.88& 91.34$\pm$0.39& 87.79$\pm$0.89& 87.50$\pm$1.33& 87.86$\pm$0.53& 83.29$\pm$1.78 \\ 
    \bottomrule
  \end{tabular}
  \end{adjustbox}
  \caption{Ablation study results (average and standard deviation of 5 trials) on CIFAR-10. }
  \label{tab:ablation}
\end{table*}

\vspace*{0.4em}
\noindent\textbf{Implementation details.}
We choose different models and number of clients $N$ for each dataset; see \cref{tab:dataset}. For data pre-processing, we perform normalization and image augmentation using random horizontal flipping and random cropping with padding=4. We use an SGD local optimizer with a momentum of 0.5, with a batch size of 10 for CIFAR-10/100 and 16 for Clothing1M. With the exception of \texttt{JointOpt} and \texttt{DivideMix} used in FL settings, we shall always use 5 local epochs across all experiments. For \texttt{FedCorr}, we always use the same hyperparameters on the same dataset. In particular, we use $T_1=5, 10, 2$ for CIFAR-10, CIFAR-100, Clothing1M, respectively. For fraction scheduling, we use the fraction $\gamma = \frac{1}{N}$ in the pre-processing stage, and we use the fractions specified in \cref{tab:dataset} for the latter two stages. Further implementation details can be found in the supplementary material; see \cref{sec:implement}. 

\begin{figure}[ht!]
  \centering
  \includegraphics[width=1.0\columnwidth]{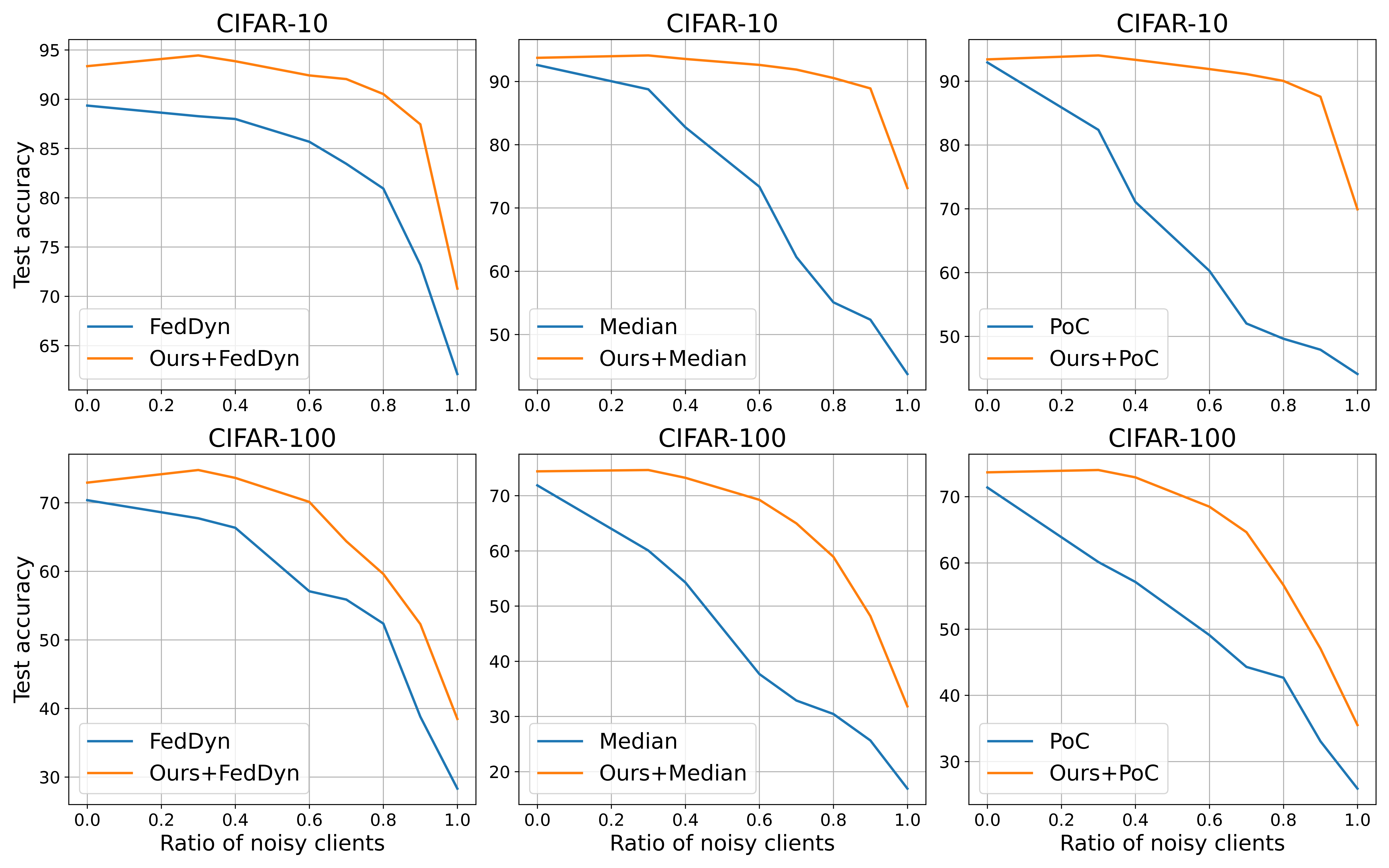}
   \caption{Best test accuracies of three FL methods combined with \texttt{FedCorr} on CIFAR-10/100 with multiple $\rho$ and fixed $\tau=0.5$.}
   \label{fig:oursplus}
\end{figure}

\subsection{Comparison with state-of-the-art methods}
\label{subsec: exp}
\noindent\textbf{IID settings}.
We compare \texttt{FedCorr} with multiple baselines at different noise levels, using the same configuration. \cref{experiment: cifar-10 IID} and \cref{experiment: cifar-100} show the results on CIFAR-10 and CIFAR-100, respectively. In summary, \texttt{FedCorr} achieves best test accuracies across all noise settings tested on both datasets, with particularly significant outperformance in the case of high noise levels. Note that we have implemented \texttt{JointOpt} and \texttt{DivideMix} in both centralized and federated settings to show the performance reduction ($10\%\sim 30\%$ lower for best accuracy) when these CL methods are applied to local clients in FL. Furthermore, the accuracies in CL can also be regarded as upper bounds for the accuracies in FL. Remarkably, the accuracy gap between \texttt{DivideMix} in CL and \texttt{FedCorr} in FL is $<4\%$ even in the extreme noise setting $(\rho, \tau)=(0.8, 0.5)$. In the centralized setting, we use the dataset corrupted with exactly the same scheme as in the federated setting. For the federated setting, we warm up the global model for 20 rounds with \texttt{FedAvg} to avoid introducing additional label noise during the correction process in the early training stage, and we then apply \texttt{JointOpt} or \texttt{DivideMix} locally on each selected client, using 20 local training epochs. 

\noindent\textbf{Non-IID settings}.
To evaluate \texttt{FedCorr} in more realistic heterogeneous data settings, we conduct experiments using the non-IID settings as described in \cref{notation and problem stagements}, over different values for $(p,\alpha_{Dir})$. \cref{tab:cifar10_noniid} and \cref{table: clothing_noniid} show the results on CIFAR-10 and Clothing1M, respectively. Note that we do not add synthetic label noise to Clothing1M, since it already contains real-world label noise. For CIFAR-10, \texttt{FedCorr} consistently outperforms all baselines by at least $7\%$. For Clothing1M, \texttt{FedCorr} also achieves the highest accuracy in FL, and this accuracy is even higher than the reported accuracy of \texttt{JointOpt} in CL. 

\noindent\textbf{Combination with other FL methods}.
We also investigate the performance of three state-of-the-art methods, when the first two stages of \texttt{FedCorr} are incorporated. As shown in \cref{fig:oursplus}, we consistently obtain significant accuracy improvements on CIFAR-10/100 for various ratios of noisy clients. 

\subsection{Ablation study}
\cref{tab:ablation} gives an overview of the effects of the components in \texttt{FedCorr}. Below, we consolidate some insights into what makes \texttt{FedCorr} successful: 
\begin{itemize}
    \item All components help to improve accuracy.
    \item Fraction scheduling has the largest effect. The small fraction used in the pre-processing stage helps to capture local data characteristics, as it avoids information loss brought by aggregation over multiple models.
    \item The highest accuracy among different noise levels is primarily achieved at a low noise level (e.g. $\rho=0.4$) and not at the zero noise level, since additional label noise could be introduced during label correction.
\end{itemize}

\section{Conclusion}
\label{sec: conclusion}
We present \texttt{FedCorr}, a general FL framework that jointly tackles the discrepancies in both local label quality and data statistics, and that performs privacy-preserving label correction for identified noisy clients. Our experiments demonstrate the robustness and outperformance of \texttt{FedCorr} at multiple noise levels and diverse data settings. 

In its current formulation, \texttt{FedCorr} does not consider dynamic participation in FL, whereby clients can join or leave training at any time. New clients joining much later would always have relatively lower cumulative LID scores, which means new noisy clients could be categorized incorrectly as clean clients. Thus, further work is required to handle dynamic participation.

{\small
\bibliographystyle{ieee_fullname}
\bibliography{egbib}
}

\begin{table*}[t!]
\centering
\begin{adjustbox}{width=\textwidth,center}
\parbox{\textwidth}{\centering\Large \textbf{Supplementary Material for\\ FedCorr: Multi-Stage Federated Learning for  Label Noise Correction}}
\end{adjustbox}
\end{table*}

\newpage

\appendix

\section{Outline}
\label{sec:outline}
As part of supplementary material for our paper titled ``FedCorr: Multi-Stage Federated Learning for Label Noise Correction'', we provide further details, organized into the following sections:
\setlist{nosep}
\begin{itemize}
    \item \cref{sec:implement} introduces the implementation details for our method and baselines.
    \item \cref{Apndxsec: exp} provides further details on our experiments. 
    \begin{itemize}
        \item \cref{subsec:100niid} gives additional experiment results on CIFAR-100 with a non-IID data partition.
        \item \cref{subsec:comp_arch} shows that \texttt{FedCorr} is model-agnostic, via a comparison of the test accuracies and the distributions of cumulative LID scores, using different model architectures.
        \item \cref{subsec:commcost} gives a comparison of the communication efficiency of different methods.
        \item \cref{subsec:lid} explains why cumulative LID scores are preferred over LID scores for identifying noisy clients.
        \item \cref{subsec:correction} demonstrates the effectiveness of both the label noise identification and the label correction process in \texttt{FedCorr}.
        \item \cref{subsec:ablation} gives further details on the ablation study results for \texttt{FedCorr}.
        \item \cref{subsec:noniid_all} provides further intuition on the non-IID data settings used in our experiments, via explicit illustrations of the corresponding non-IID data partitions on CIFAR-10, over 100 clients.
    \end{itemize}
    \item \cref{sec:neg_impact} discusses the potential negative societal impact of \texttt{FedCorr}.
\end{itemize}

\section{Implementation details}
\label{sec:implement}
\begin{table}[ht!]
  \centering
  \begin{adjustbox}{width=\columnwidth,center}
  \begin{tabular}{lccc}
    \toprule
    \midrule
    Hyperparameters & CIFAR-10   & CIFAR-100  & Clothing1M \\
    \midrule
    $\#$ of iterations in stage 1, $T_1$    & 5   & 10  & 2   \\
    $\#$ of rounds in stage 2, $T_2$  & 500  & 450  & 50  \\
    $\#$ of rounds in stage 3, $T_3$  & 450  & 450  & 50  \\
    Confidence threshold, $\theta$  & 0.5 & 0.5 & 0.9  \\
    Relabel ratio, $\pi$   & 0.5 & 0.5 & 0.8  \\
    Learning rate   & 0.03  & 0.01  & 0.001  \\   
    \bottomrule
\end{tabular}
\end{adjustbox}
\caption{Hyperparameters of \texttt{FedCorr} on different datasets.}
\label{tab: hyperparameters}
\end{table}

All experiments were implemented using Pytorch. 
Among the baselines, we reimplemented \texttt{FedAvg}~\cite{mcmahan2017communication}, \texttt{FedProx}~\cite{li2020prox}, \texttt{JointOpt}~\cite{tanaka2018joint}, \texttt{DivideMix}~\cite{Li2020DivideMix} and \texttt{PoC}~\cite{cho2020client}, and we used the official implementations of \texttt{FedDyn}~\cite{acar2020federated} and  \texttt{ARFL}~\cite{fu2019attack}. For \texttt{RoFL}\footnote{https://github.com/jangsoohyuk/Robust-Federated-Learning-with-Noisy-Labels} and \texttt{Median}\footnote{https://github.com/fushuhao6/Attack-Resistant-Federated-Learning}, we used their unofficial implementations. For all methods, we use an SGD local optimizer with a momentum of 0.5 and no weight decay, with a batch size of 10 for CIFAR-10/100 and 16 for Clothing1M. Note that at each noise level, we used the same training hyperparameters for both IID and non-IID data partitions. 

For the implementation of each federated learning (FL) method, we define its \textit{total communication cost} to be the cumulative number of clients that participate in training. For example, if a client participates in 10 communication rounds, then that client would contribute 10 to the total communication cost. For every method except \texttt{JointOpt} and \texttt{DivideMix}, we always reimplement the method using 5 local epochs per communication round, and the same total communication cost for each dataset, which corresponds to 1000 rounds of \texttt{FedAvg} for CIFAR-10/100 with fraction 0.1, and corresponds to 200 rounds of \texttt{FedAvg} for Clothing1M with fraction 0.02. Settings for \texttt{JointOpt} and \texttt{DivideMix} are discussed below.

In the rest of this section, we give full details on all remaining hyperparameters used for each method. 
For baseline methods, we also provide brief descriptions of their main underlying ideas.
\begin{itemize}
    \item \texttt{FedCorr.} We fixed $k=20$ for LID estimation, $\alpha=1$ for mixup, and $\beta=5$ for the proximal regularization term in all reported experiments. All remaining hyperparameters can be found in \cref{tab: hyperparameters}. Note that the total communication cost for \texttt{FedCorr} is the same as for other baselines. 
    Take CIFAR-10 for example: In each iteration of the pre-processing stage of \texttt{FedCorr}, every client participates exactly once. In contrast, in each communication round of our other baselines, only a fraction of 0.1 of the clients would participate.
    Hence, one iteration in the pre-processing stage of \texttt{FedCorr} has 10 times the total communication cost of one communication round of the other baselines.

    For the latter two stages of \texttt{FedCorr}, we used the usual 0.1 as our fraction.
    Hence the total communication cost of the entire implementation of \texttt{FedCorr} equals $100T_1+10T_2+10T_3=10000$; this is the same total communication cost for implementing \texttt{FedAvg} over $1000$ communication rounds with fraction 0.1.
    \item \texttt{JointOpt} \cite{tanaka2018joint} is one of the state-of-the-art centralized methods for tackling label noise, which alternately updates network parameters and corrects labels using the model prediction vectors. It introduced $\alpha_{Jo}$ and $\beta_{Jo}$ as two additional hyperparameters. In the centralized setting, we used the hyperparameters given in \cref{tab: hyperparameters_Jo2}. In particular, we considered a total of seven noise settings, which we have divided into two groups: low noise levels (first four settings) and high noise levels (last three settings). Within each group, we used the same hyperparameters. Note that the hyperparameters are not exactly the same as those given in \cite{tanaka2018joint}, as we used different architectures and different frameworks to generate synthetic label noise. In the federated setting, we used $\alpha_{Jo}=1.2$, $\beta_{Jo}=0.8$ and a learning rate of $0.01$ for CIFAR-10/100. To boost performance, we used a warm-up process for CIFAR-10/100: We first trained using \texttt{FedAvg} over 20 communication rounds with 5 local epochs per communication round, after which we started using \texttt{JointOpt} for local training over 80 communication rounds with 20 local epochs per communication round. For Clothing1M, we used $\alpha_{Jo}=1.2$, $\beta_{Jo}=0.8$, and a learning rate of $0.001$. As we used a ResNet-50 that is already pretrained on ImageNet, no warm-up process was used for our Clothing1M experiments. We trained using \texttt{JointOpt} over 40 communication rounds with 10 local epochs per round.

    \begin{table}[ht!]
    \begin{adjustbox}{width=\columnwidth,center}
    \begin{tabular}{lccc}
        \toprule
        \midrule
        $(\rho,\tau)$  & $\alpha_{Jo}$ & $\beta_{Jo}$ & Learning rate \\
        \midrule
Low noise levels & \multirow{2}{*}{1.2} & \multirow{2}{*}{1.5} & \multirow{2}{*}{0.1} \\
        $(0.0,0.0),(0.4,0.0),(0.4,0.5),(0.6,0.0)$  &  &  &  \\
        \midrule
High noise levels & \multirow{2}{*}{1.2} & \multirow{2}{*}{0.8} & \multirow{2}{*}{0.2} \\
         $(0.6,0.5),(0.8,0.0),(0.8,0.5)$ &  &  & \\
        \bottomrule
    \end{tabular}
    \end{adjustbox}
    \caption{Hyperparameters of \texttt{JointOpt} in the centralized setting on CIFAR-10/100.}
    \label{tab: hyperparameters_Jo2}
    \end{table}
    
    \item \texttt{DivideMix} \cite{Li2020DivideMix} is another state-of-the-art centralized method, which dynamically divides the training data into labeled (clean) and unlabeled (noisy) data, and trains the model in a semi-supervised manner. For CIFAR-10/100, we used the same two groups of noise settings, as described in the above configuration for \texttt{JointOpt}. The only hyperparameter we tuned is $\lambda_{Div}$, which is a hyperparameter specific to \texttt{DivideMix}. For low noise levels, we used $\lambda_{Div}=0$ (resp. $\lambda_{Div}=25$) for CIFAR-10 (resp. CIFAR-100). For high noise levels, we used $\lambda_{Div}=25$ (resp. $\lambda_{Div}=150$) for CIFAR-10 (resp. CIFAR-100). For all other hyperparameters for CIFAR-10/100, we used the values given in \cite{Li2020DivideMix}. For Clothing1M, we use $\lambda_{Div}=25$ and a learning rate of $0.01$; for all other hyperparameters, we used the values given in \cite{Li2020DivideMix}. We used the same warm-up process for CIFAR-10/100, and we used the same number of communication rounds and number of local epochs for all datasets, as described above in our configuration for \texttt{JointOpt}. 
    
    \item \texttt{FedAvg} \cite{mcmahan2017communication} is the first algorithm that introduced the idea of federated learning. We used a learning rate of 0.03, 0.01 and 0.003 for CIFAR-10, CIFAR-100 and Clothing1M, respectively.
    
    \item \texttt{FedProx} \cite{li2020prox} was proposed to tackle data heterogeneity among clients by adding a fixed proximal term with coefficient $\mu_{prox}$ to every local loss function. We used $\mu_{prox}=1$ for all experiments, and a learning rate of 0.01 and 0.003 for CIFAR-10/100 and Clothing1M, respectively.
    
    \item \texttt{RoFL} \cite{yang2020robust} is, to the best of our knowledge, the only method that is designed for label correction in FL. It is based on the idea of exchanging feature centroids between the server and clients, and it introduced $T_{pl}$ as an additional hyperparameter to control label correction. We set $T_{pl}$ to 100, 400 and 10 for CIFAR-10, CIFAR-100 and Clothing1M, respectively. All other hyperparameters are set to the same values as given in \cite{yang2020robust}.
    
    \item \texttt{ARFL} \cite{fu2019attack} is a robust aggregation algorithm that resists abnormal attacks via residual-based reweighting, using two hyperparameters $\lambda_{ar}$ and threshold $\delta_{ar}$. We used $\lambda_{ar}=2$ and $\delta_{ar}=0.1$ for all experiments. We used a learning rate of $0.01$ and $0.003$ for CIFAR-10/100 and Clothing1M, respectively. 
    
    \item \texttt{FedDyn} \cite{acar2020federated} proposed a dynamic regularizer, with coefficient $\alpha_{Dyn}$, for local optimization in each communication round, so as to tackle the inconsistency between the local and global empirical loss. We used $\alpha_{Dyn}=0.01$, a learning rate of 0.1 with a decay of 0.998 for all the experiments.
    
    \item \texttt{Median} \cite{pmlr-v80-yin18a} is an aggregation method for robust distributed learning, whereby the notion of ``average'' in \texttt{FedAvg} is changed from ``mean'' to ``median''. For all experiments, we used a learning rate of 0.01; all other hyperparameters are the same as given in \texttt{FedAvg}. 
    
    \item \texttt{Poc} \cite{cho2020client} is a client selection algorithm that is biased towards clients with higher local losses within a given client pool. We used a learning rate of 0.01 and a client pool size of $d=30$ for all experiments.
\end{itemize}

\section{Details on experiment results}
\label{Apndxsec: exp}

\begin{table}[ht!]
  \centering
  \begin{tabular}{lc}
    \toprule
    \midrule
    Method$\backslash$($p,\alpha_{Dir}$) & $(0.7,10)$ \\
    \midrule
    FedAvg & 64.75$\pm$1.75 \\
    FedProx &65.72$\pm$1.30  \\
    RoFL  &59.31$\pm$4.14 \\
    ARFL  &48.03$\pm$4.39\\
    JointOpt &59.84$\pm$1.99  \\
    DivideMix &39.76$\pm$1.18 \\
    Ours &\textbf{72.73$\pm$1.02}  \\
    \bottomrule
  \end{tabular}
  \caption{Average (5 trials) and standard deviation of the best test accuracies of different methods on CIFAR-100 with non-IID data partition. The noise setting used is $(\rho,\tau)=(0.4,0)$.}
  \label{tab:cifar100_noniid}
\end{table}

\begin{table*}[htb!]
  \centering
  \begin{tabular}{l|ccccccc}
    \toprule
    \midrule
    \multirow{3}{*}{Method} & \multicolumn{7}{|c}{Best Test Accuracy (\%) $\pm$ Standard Deviation (\%)} \\
    \cmidrule{2-8}
     & \multicolumn{1}{|c|}{$\rho=0.0$} & \multicolumn{2}{|c|}{$\rho=0.4$} & \multicolumn{2}{c}{$\rho=0.6$} & \multicolumn{2}{|c}{$\rho=0.8$} \\ \cmidrule{2-8} 
     & \multicolumn{1}{c|}{$\tau=0.0$} & \multicolumn{1}{c|}{$\tau=0.0$} & \multicolumn{1}{c|}{$\tau=0.5$} & \multicolumn{1}{c|}{$\tau=0.0$} & \multicolumn{1}{c|}{$\tau=0.5$} & \multicolumn{1}{c|}{$\tau=0.0$} & \multicolumn{1}{|c}{$\tau=0.5$} \\
    \midrule
      \multicolumn{1}{l|}{ResNet-18} & 93.82$\pm$0.41& 94.01$\pm$0.22& 94.15$\pm$0.18& 92.93$\pm$0.25& 92.50$\pm$0.28& 91.52$\pm$0.50& 90.59$\pm$0.70 \\
      \multicolumn{1}{l|}{VGG-11} &88.96$\pm$0.84&87.93$\pm$0.41&87.53$\pm$0.40&84.78$\pm$1.68&84.82$\pm$0.79&83.34$\pm$0.42&80.82$\pm$2.62  \\
      \multicolumn{1}{l|}{LeNet-5} & 72.03$\pm$0.35& 70.47$\pm$0.86& 70.02$\pm$1.39& 69.09$\pm$0.16& 67.48$\pm$0.54& 67.49$\pm$0.74& 65.16$\pm$0.53 \\
    \bottomrule
  \end{tabular}
  \caption{Comparison of the average (5 trials) and standard deviation of best test accuracies, when trained on CIFAR-10 with IID data partition using different architectures.}
  \label{tab:acrhi}
\end{table*}

\begin{figure*}[htb!]
  \centering
  \includegraphics [width=1.0\linewidth]
  {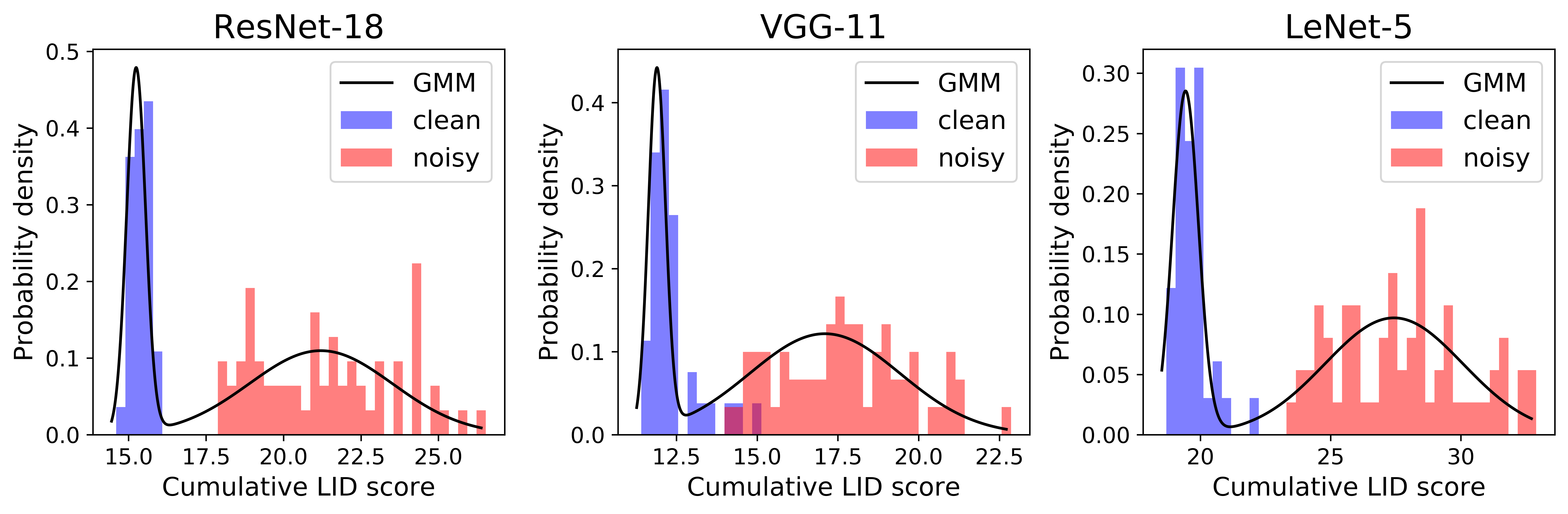}
   \caption{Comparison of cumulative LID score distribution after 5 iterations in pre-processing stage among different architectures. The experiments were conducted on CIFAR-10 with IID data partition and noise setting $(\rho,\tau)=(0.6,0.5)$.}
   \label{fig:lid_models}
\end{figure*}

\subsection{CIFAR-100 with non-IID data partition}
\label{subsec:100niid}

\begin{table*}[htb!]
  \centering
  \begin{tabular}{l|rrrrrrr}
    \toprule
    \midrule
    \multirow{2}{*}{Method}  & \multicolumn{1}{|c|}{$\rho=0.0$} & \multicolumn{2}{|c|}{$\rho=0.4$} & \multicolumn{2}{c}{$\rho=0.6$} & \multicolumn{2}{|c}{$\rho=0.8$} \\ \cmidrule{2-8} 
     & \multicolumn{1}{c|}{$\tau=0.0$} & \multicolumn{1}{c|}{$\tau=0.0$} & \multicolumn{1}{c|}{$\tau=0.5$} & \multicolumn{1}{c|}{$\tau=0.0$} & \multicolumn{1}{c|}{$\tau=0.5$} & \multicolumn{1}{c|}{$\tau=0.0$} & \multicolumn{1}{|c}{$\tau=0.5$} \\
    \midrule
      \multicolumn{1}{l|}{Ours} &\multicolumn{1}{l}{150}&\multicolumn{1}{l}{\;\,210}&\multicolumn{1}{l}{\;\,230}&\multicolumn{1}{l}{  \;\,230}&\multicolumn{1}{l}{\quad330}&\multicolumn{1}{l}{  \;\,360}&\multicolumn{1}{l}{\;\,510} \\
      \midrule
      \multicolumn{1}{l|}{FedAvg}  &370(2.6$\times$)&450(2.1$\times$)&470(2.0$\times$)&550(2.4$\times$)&930(2.8$\times$)&810(2.3$\times$)&-  \\
      \multicolumn{1}{l|}{FedProx}  &690(4.9$\times$)&1050(5.0$\times$)&1190(5.2$\times$)&1230(5.3$\times$)&1600(4.8$\times$)&1730(4.8$\times$)&4640(9.1$\times$) \\
      \multicolumn{1}{l|}{RoFL}   &990(7.1$\times$)&1390(6.6$\times$)&1580(6.9$\times$)&1900(8.3$\times$)&4200(12.7$\times$)&2080(5.8$\times$)&- \\
      \multicolumn{1}{l|}{ARFL} &290(2.1$\times$)&740(3.5$\times$)&1180(5.1$\times$)&-&-&-&- \\
      \multicolumn{1}{l|}{JointOpt} &330(2.4$\times$)&420(2.0$\times$)&760(3.3$\times$)&550(2.4$\times$)&-&-&- \\ 
      \multicolumn{1}{l|}{DivideMix} &-&-&-&-&-&-&- \\ 
    \bottomrule
  \end{tabular}
  \caption{A comparison of communication efficiency for different methods on CIFAR-10 with IID data partition, in terms of the targeted communication cost at $\zeta=80\%$ test accuracy. Values in brackets represent the ratios of the targeted communication costs as compared to our method \texttt{FedCorr}. Note that the test accuracies are evaluated after each communication round. In the case of methods and noise settings for which the target test accuracy $\zeta$ is not reached, we indicate `-'.}
  \label{tab:cost 80 cifar10}
\end{table*}

In terms of robustness to the discrepancies in both local label quality and local data statistics, \texttt{FedCorr} significantly outperforms the baselines. In the main paper, we have reported the outperformance of \texttt{FedCorr} on CIFAR-100 with IID data partition. To further show the outperformance on non-IID data partitions, we also conducted experiments on CIFAR-100 with noise model $(\rho, \tau)=(0.4,0.5)$ and non-IID hyperparameter $(p, \alpha_{Dir})=(0.7,10)$; here, we report our results in \cref{tab:cifar100_noniid}. We observe that \texttt{FedCorr} achieves an improvement in best test accuracy of at least $7\%$ over our baselines.

\begin{table*}[htb!]
  \centering
  \begin{tabular}{l|rrrrrrr}
    \toprule
    \midrule
    \multirow{2}{*}{Method}  & \multicolumn{1}{|c|}{$\rho=0.0$} & \multicolumn{2}{|c|}{$\rho=0.4$} & \multicolumn{2}{c}{$\rho=0.6$} & \multicolumn{2}{|c}{$\rho=0.8$} \\ \cmidrule{2-8} 
     & \multicolumn{1}{c|}{$\tau=0.0$} & \multicolumn{1}{c|}{$\tau=0.0$} & \multicolumn{1}{c|}{$\tau=0.5$} & \multicolumn{1}{c|}{$\tau=0.0$} & \multicolumn{1}{c|}{$\tau=0.5$} & \multicolumn{1}{c|}{$\tau=0.0$} & \multicolumn{1}{|c}{$\tau=0.5$} \\
    \midrule
      \multicolumn{1}{l|}{Ours} &\multicolumn{1}{l}{\;\,50}&\multicolumn{1}{l}{\;\,60}&\multicolumn{1}{l}{\;\,90}&\multicolumn{1}{l}{\;\,70}&\multicolumn{1}{l}{110}&\multicolumn{1}{l}{\quad90}&\multicolumn{1}{l}{\;\,190} \\
      \midrule
      \multicolumn{1}{l|}{FedAvg}  &160(3.2$\times$)&200(3.3$\times$)&210(2.3$\times$)&230(3.3$\times$)&300(2.7$\times$)&270(3.0$\times$)&470(2.5$\times$)  \\
      \multicolumn{1}{l|}{FedProx}  &300(6.0$\times$)&430(7.2$\times$)&500(5.6$\times$)&480(6.9$\times$)&690(6.3$\times$)&670(7.4$\times$)&1840(9.7$\times$) \\
      \multicolumn{1}{l|}{RoFL}   &350(7.0$\times$)&420(7.0$\times$)&470(5.2$\times$)&440(6.3$\times$)&670(6.1$\times$)&490(5.4$\times$)&1710(9.0$\times$) \\
      \multicolumn{1}{l|}{ARFL} &120(2.4$\times$)&230(3.8$\times$)&170(1.9$\times$)&240(3.4$\times$)&390(3.5$\times$)&270(3.0$\times$)&- \\
      \multicolumn{1}{l|}{JointOpt} &160(3.2$\times$)&200(3.3$\times$)&220(2.4$\times$)&220(3.1$\times$)&250(2.3$\times$)&250(2.8$\times$)&860(4.5$\times$) \\ 
      \multicolumn{1}{l|}{DivideMix} &480(9.6$\times$)&560(9.3$\times$)&580(6.4$\times$)&590(8.4$\times$)&690(6.3$\times$)&930(10.3$\times$)&970(5.1$\times$) \\ 
    \bottomrule
  \end{tabular}
  \caption{A comparison of communication efficiency for different methods on CIFAR-10 with IID data partition, in terms of the targeted communication cost at $\zeta=65\%$ test accuracy. Values in brackets represent the ratios of the targeted communication costs as compared to our method \texttt{FedCorr}. Note that the test accuracies are evaluated after each communication round. Note that the test accuracies are evaluated after each communication round. In the case of methods and noise settings for which the target test accuracy $\zeta$ is not reached, we indicate `-'.}
  \label{tab:cost 65 cifar10}
\end{table*}

\begin{table}[htb!]
  \centering
  \begin{adjustbox}{width=\columnwidth,center}
  \begin{tabular}{l|rrrr}
    \toprule
    \midrule
     \multirow{2}{*}{Method}  &\multicolumn{1}{c|}{$\rho=0.0$}  &\multicolumn{1}{c|}{$\rho=0.4$} &\multicolumn{1}{c|}{$\rho=0.6$}  &\multicolumn{1}{c}{$\rho=0.8$}\\
    &\multicolumn{1}{c|}{$\tau=0.0$}  &\multicolumn{1}{c|}{$\tau=0.5$} &\multicolumn{1}{c|}{$\tau=0.5$} &\multicolumn{1}{c}{$\tau=0.5$}\\
    \midrule
      Ours  &\multicolumn{1}{l}{\,\;95}&\multicolumn{1}{l}{\quad140} &\multicolumn{1}{l}{\quad295} &\multicolumn{1}{l}{\,\;505} \\ 
     \midrule
      FedAvg     &135(1.4$\times$) &210(1.5$\times$)   &420(1.4$\times$)  &- \\
      FedProx    &465(4.9$\times$) &705(5.0$\times$) &1110(3.8$\times$) &1885(3.7$\times$) \\
      RoFL  &380(4.0$\times$) &2350(16.8$\times$) &4740(16.1$\times$) &- \\
      ARFL  &150(1.6$\times$) &265(1.9$\times$)   &- &- \\
      JointOpt   &125(1.3$\times$) &210(1.5$\times$)   &- &-  \\
      DivideMix  &- &- &- &-  \\
    \bottomrule
  \end{tabular}
  \end{adjustbox}
  \caption{A comparison of communication efficiency for different methods on CIFAR-100 with IID data partition, in terms of the targeted communication cost at $\zeta=50\%$ test accuracy. Values in brackets represent the ratios of the targeted communication costs as compared to our method \texttt{FedCorr}. Note that the test accuracies are evaluated after each communication round. Note that the test accuracies are evaluated after each communication round. In the case of methods and noise settings for which the target test accuracy $\zeta$ is not reached, we indicate `-'.}
  \label{tab:cost cifar-100}
\end{table}

\subsection{Comparison of different architectures}
\label{subsec:comp_arch}

To demonstrate that our proposed \texttt{FedCorr} is model-agnostic, especially with respect to the noisy client identification scheme via cumulative LID scores, we conducted experiments on CIFAR-10 with IID data partition using different architectures: ResNet-18 \cite{he2016deep}, VGG-11 \cite{SimonyanZ14a} and LeNet-5 \cite{lecun1989backpropagation}. \cref{tab:acrhi} shows the best test accuracies of each model trained on CIFAR-10 with various levels of synthetic noise. For experiments on VGG-11, we used hyperparameters with the same values as used in the experiments on ResNet-18. For LeNet-5, we only tuned the learning rate and fixed it at 0.003 in all experiments. \cref{fig:lid_models} shows a further comparison between different architectures in terms of the distribution of the cumulative LID scores and the corresponding separations of the clients via Gaussian Mixture Models.

\subsection{Comparison of communication efficiency}
\label{subsec:commcost}

In this subsection, we discuss the communication efficiency of different methods. Here, given any implementation of an FL method, and any desired target accuracy $\zeta$, we define its \textit{targeted communication cost for $\zeta$ test accuracy} to be the lowest total communication cost required (in the experiments) to reach the target $\zeta$ test accuracy. Informally, the lower the targeted communication cost, the higher the communication efficiency.

\cref{tab:cost 80 cifar10} and \cref{tab:cost 65 cifar10} show the comparison of the communication efficiency on CIFAR-10, in terms of the targeted communication cost at test accuracies $\zeta=80\%$ and $\zeta=65\%$, respectively. \cref{tab:cost cifar-100} shows the comparison on CIFAR-100, in terms of the targeted communication cost at test accuracy $\zeta=50\%$. As our results show, \texttt{FedCorr} achieves improvements in communication efficiency, by a factor of at least 1.9 on CIFAR-10, and at least 1.3 on CIFAR-100.

\begin{figure*}[ht!]
  \centering
  \includegraphics [width=1.0\linewidth]
  {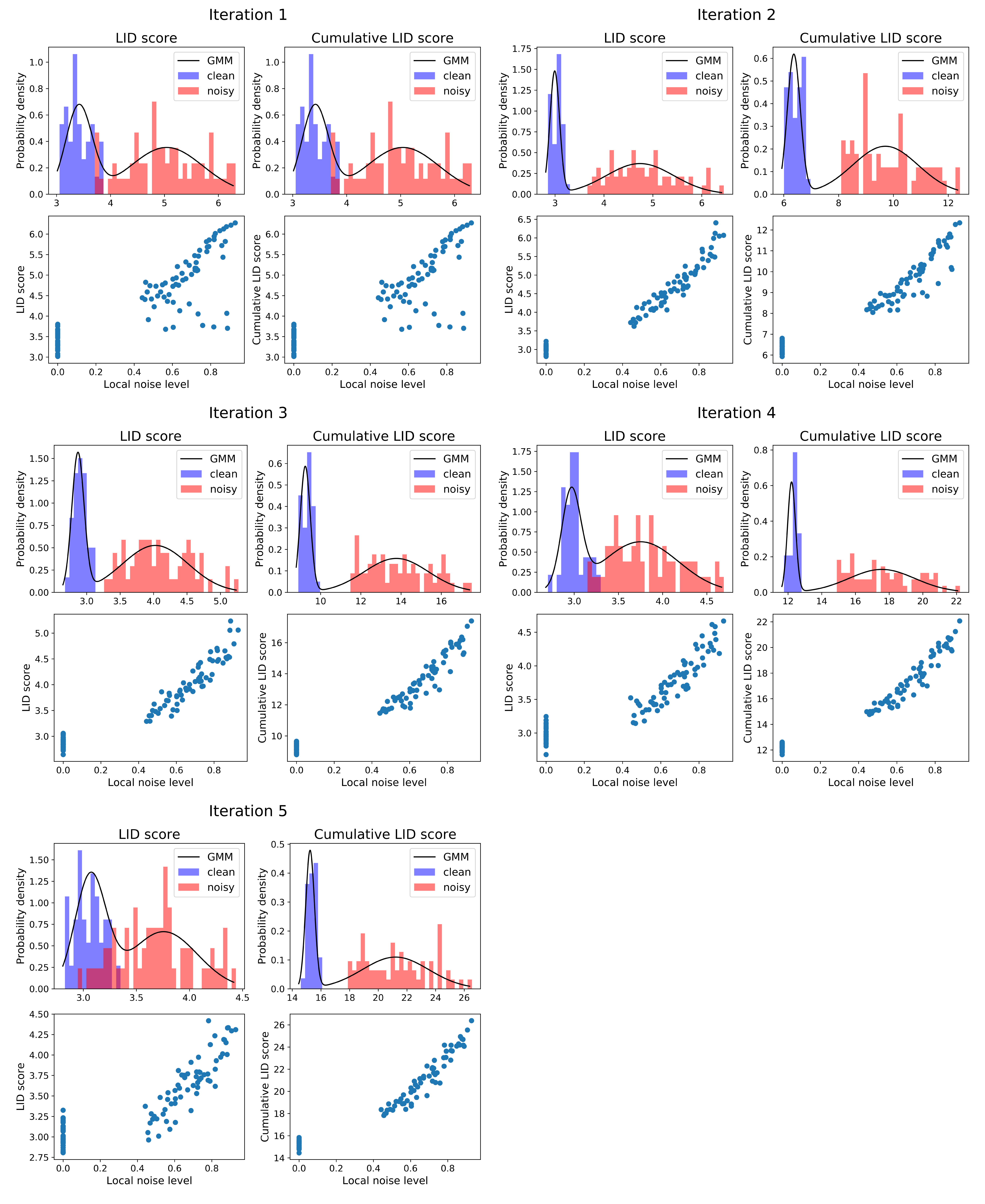}
   \caption{Distributions of the LID/cumulative LID scores during all 5 iterations of the pre-processing stage of \texttt{FedCorr}, evaluated on CIFAR-10 with IID data partition and noise setting $(\rho, \tau)=(0.6,0.5)$, over 100 clients.}
   \label{fig:lid}
\end{figure*}

\begin{figure*}[ht!]
  \centering
  \includegraphics [width=\linewidth]
  {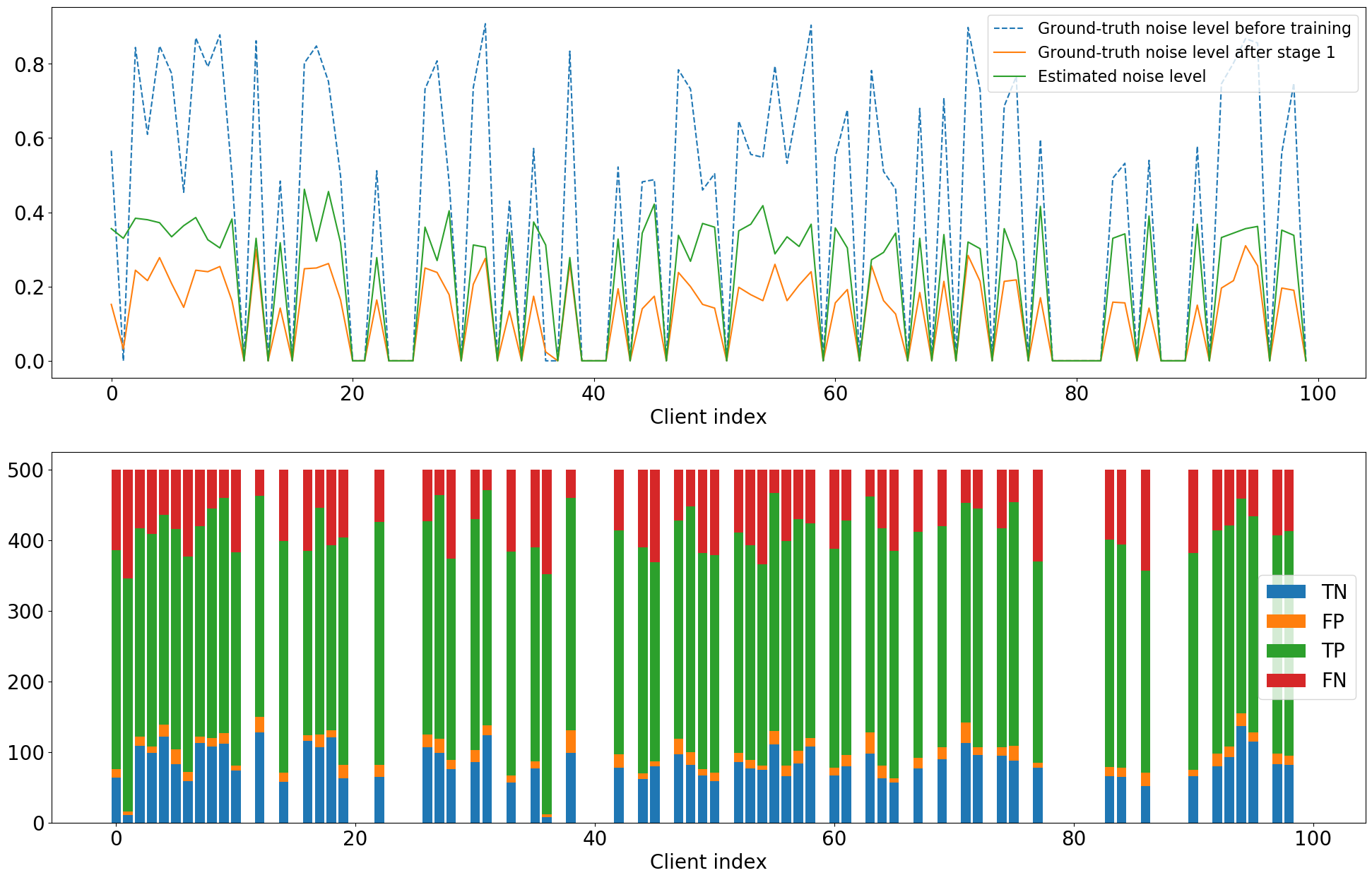}
   \caption{An evaluation of the label noise identification and label correction process after 5 iterations in the pre-processing stage, conducted on CIFAR-10 with IID data partition and noise setting $(\rho, \tau)=(0.6,0.5)$. Top: Evaluation of noise level estimation and label correction process in the pre-processing stage. Bottom: Evaluation of label noise identification.}
   \label{fig:estimation}
\end{figure*}

\subsection{Distribution of cumulative LID scores}
\label{subsec:lid}
\cref{fig:lid} shows the comparison between the distribution of the LID scores and the distribution of the cumulative LID scores, after each iteration in the pre-processing stage. The LID scores of clean clients and noisy clients can be well-separated after the second iteration and the third iteration. This is also true for the cumulative LID scores. However, after the fourth iteration, the LID scores of noisy clients and clean clients start overlapping, while in contrast, the cumulative LID scores of noisy clients and clean clients remain well-separated. As already discussed in the main paper, cumulative LID scores have a stronger linear relation with local noise levels, as compared to LID scores. Hence, the cumulative LID score is a more robust metric for identifying noisy clients.

\subsection{Evaluation of label noise identification and label correction}
\label{subsec:correction}

\cref{fig:estimation} demonstrates the effectiveness of the label noise identification and correction process in the pre-processing stage on CIFAR-10. Note that in \cref{fig:estimation}, we used the noise setting $(\rho, \tau)=(0.6,0.5)$, which means on average $60\%$ of the clients are randomly selected for the addition of synthetic noise to their local datasets before training, whereby the local noise level for each selected client is at least $0.5$. The top plot in \cref{fig:estimation} shows the estimated noise levels, in comparison with the ground-truth noise levels (before training and after stage 1), across all 100 clients. In particular, the huge gap between the ground-truth noise levels before training (blue dotted line) and after stage 1 (orange line) represents the effectiveness of our label correction process, while the small gap between the estimated noise levels (green line) and the ground-truth noise levels after stage 1 (orange line) reflects the effectiveness of our local noise level estimation. Note that for clean clients (with zero ground-truth noise levels before training), \texttt{FedCorr} is able to estimate their noise levels to be exactly zero in most cases. Consequently, no additional label noise is introduced to these identified clean clients in our label correction process. 

The bottom plot in \cref{fig:estimation} shows the separation results between noisy and clean samples (via a Gaussian Mixture Model) for each identified noisy client, in terms of true/false positives/negatives. In particular, the small numbers of false positives across all identified noisy clients imply the effectiveness of \texttt{FedCorr} in identifying noisy samples.

\begin{figure*}[ht!]
  \centering
  \includegraphics [width=1.0\linewidth]
  {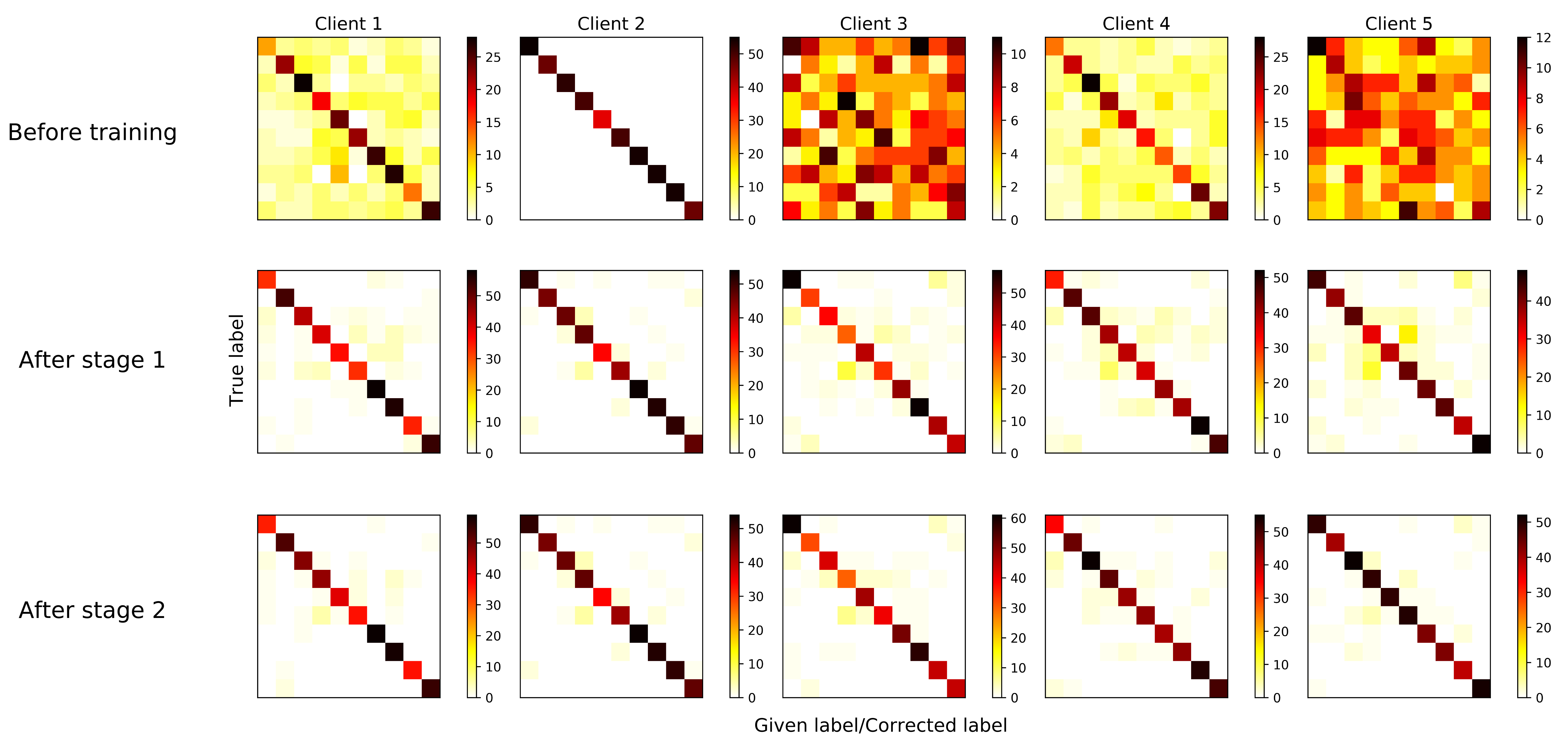}
   \caption{An evaluation of the label correction process on the first five clients, conducted on CIFAR-10 with IID data partition and noise setting $(\rho, \tau)=(0.6,0.5)$. For each client, we give the heat maps of three confusion matrices, associated to the given labels before training, the corrected labels after the pre-processing stage (stage 1), and the corrected labels after the finetuning stage (stage 2).}
   \label{fig:confusion}
\end{figure*}

To further illustrate the effectiveness of the label correction process, we compared the confusion matrices of the given labels before training, the corrected labels after the pre-processing stage, and the corrected labels after the finetuning stage. 
\cref{fig:confusion} depicts the confusion matrices for the first 5 clients, in the experiments conducted on CIFAR-10 with IID data partition and noise setting $(\rho, \tau)=(0.6,0.5)$. 
For all five selected clients, the ground-truth noise levels after label correction are close to $0$.
Notice also that for client 2, whose dataset initially has no noisy labels, only a minimal amount of label noise is introduced during the label correction process.

\begin{figure*}[htb!]
  \centering
  \includegraphics [width=0.8\linewidth]
  {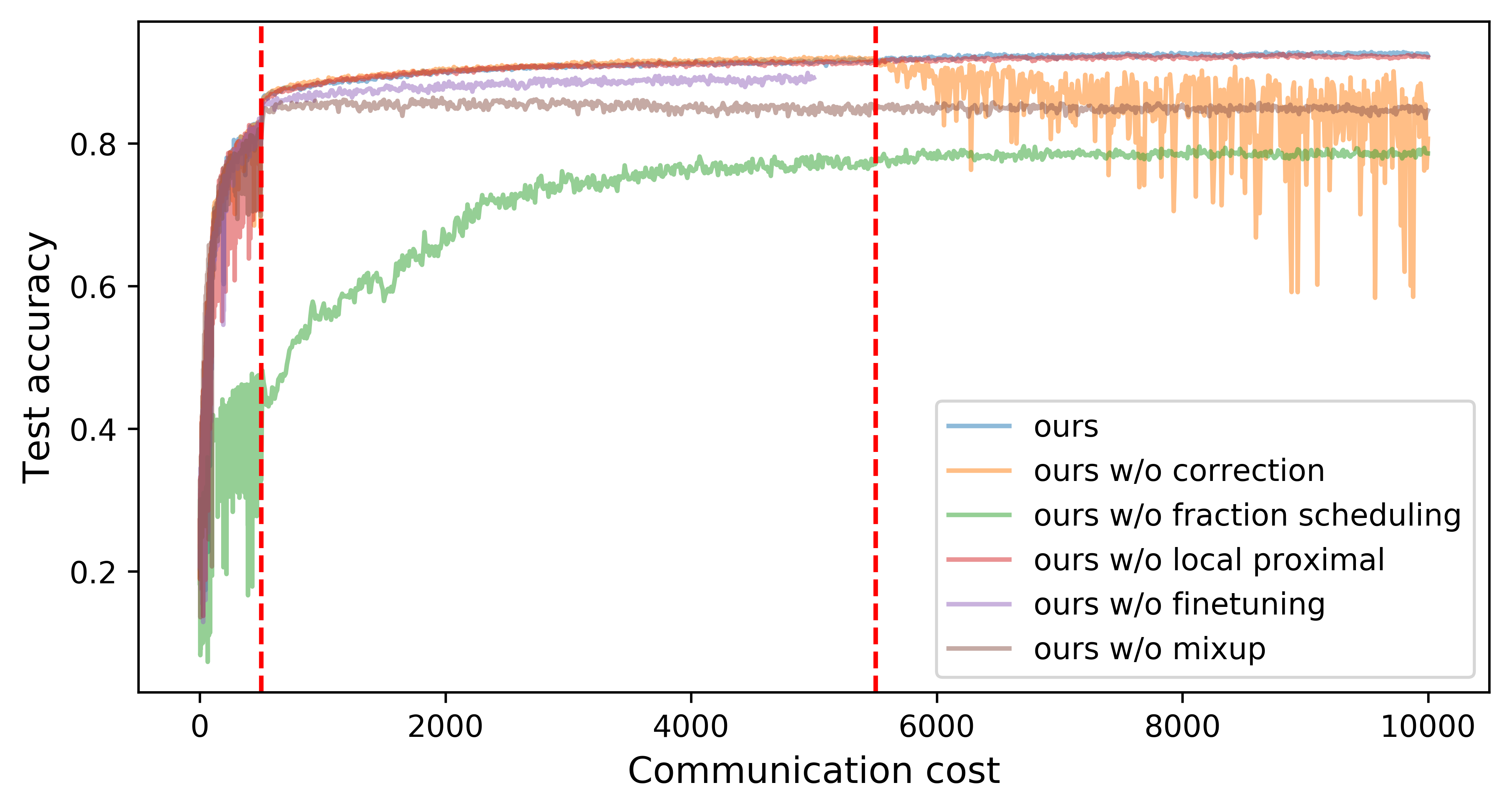}
   \caption{Ablation study results on the test accuracies of \texttt{FedCorr} during the training process, with each component removed. The experiments are evaluated on CIFAR-10 with IID data partition and noise setting $(\rho,\tau)=(0.6,0.5)$. The dotted lines represent the separation of the training process into our three stages.}
   \label{fig:ablation}
\end{figure*}

\subsection{Additional ablation study results}
\label{subsec:ablation}
In \cref{fig:ablation}, we show the effects of the components  of \texttt{FedCorr} on test accuracies during training. In particular, note that without the finetuning stage, the total communication cost would be 5000. Hence in \cref{fig:ablation}, the curve plotted for \texttt{FedCorr} without finetuning ends at the 5000 communication cost mark, which is to the left of the second red dotted line (5500 communication cost). As we mentioned in the main paper, fraction scheduling plays the most significant role in \texttt{FedCorr}. In addition, the label correction process would significantly improve training stability, especially in the usual training stage.

\begin{figure*}[htb!]
  \centering
  \includegraphics [width=1.0\linewidth]
  {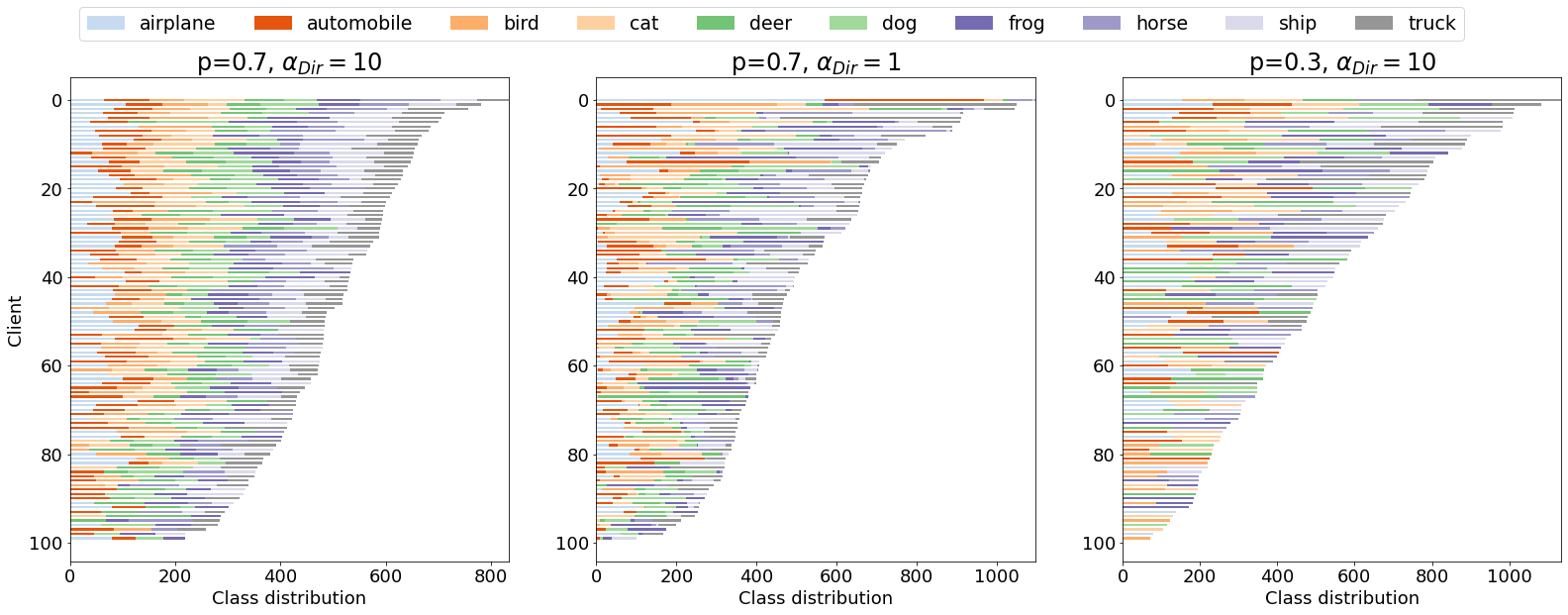}
   \caption{An illustration of three non-IID data partitions (via three different values for ($p$,$\alpha_{Dir}$)) on CIFAR-10, as used for our experiments reported in the main paper. For ease of viewing, we have sorted the clients according to their local dataset sizes in each of the three non-IID data partitions depicted. It should be noted that in our experiments, clients are \emph{not} sorted; instead, clients are assigned data samples according to the data partition described in the main paper, which is a random process with no sorting.}
   \label{fig:noniidall}
\end{figure*}

\subsection{Illustration of non-IID data partitions on CIFAR-10}
\label{subsec:noniid_all}
As reported in the main paper, we used 3 different non-IID local data settings ($(p,\alpha_{Dir})=(0.7,10), (0.7,1), (0.3,10)$) for our experiment involving non-IID data partitions. In \cref{fig:noniidall}, we illustrate the detailed local class distributions and local dataset sizes for these three non-IID data settings on CIFAR-10, over $100$ clients. 

\section{Potential negative impact: the issue of freeloaders}
\label{sec:neg_impact}
In real-world FL implementations, there is the implicit assumption that clients are collaborative and jointly collaborate to train a global model. Although \texttt{FedCorr} allows for a robust training of a global model even when some clients have label noise, this also includes the case when a client is a ``freeloader'', where the client's local dataset has completely random label noise (e.g. randomly assigning labels to an unlabeled dataset, without any actual non-trivial annotation effort). By participating in the \texttt{FedCorr} FL framework, such a ``freeloader'' would effectively use \texttt{FedCorr} as the actual annotation process, whereby identified noisy labels are corrected. Hence, this would be unfair to clients that have performed annotation on their local datasets prior to participating in \texttt{FedCorr}.

\end{document}